\documentclass[]{opendatalab}
\usepackage{xcolor}
\usepackage{longtable}
\usepackage{listings}
\usepackage{wrapfig}
\usepackage[numbers]{natbib}
\usepackage{enumitem}
\usepackage{arydshln}

\definecolor{codegreen}{rgb}{0,0.6,0}
\definecolor{codegray}{rgb}{0.5,0.5,0.5}
\definecolor{codepurple}{rgb}{0.58,0,0.82}
\definecolor{backcolour}{rgb}{0.95,0.95,0.92}
\definecolor{promptcolor}{HTML}{D1D0F2}
\definecolor{promptcolorheader}{HTML}{bdbcec}
\newcommand{\promptbox}[2]{
\begin{tcolorbox}[
top=0.3em,bottom=0.3em,left=0.5em,right=0.5em,
toptitle=0.3em,bottomtitle=0.2em,boxsep=0pt,
colframe=promptcolorheader,colback=promptcolor!50,boxrule=0.5pt,
]
\footnotesize
\end{tcolorbox}
}
\lstdefinestyle{mystyle}{
    backgroundcolor=\color{backcolour},   
    commentstyle=\color{codegreen},
    keywordstyle=\color{magenta},
    numberstyle=\tiny\color{codegray},
    stringstyle=\color{codepurple},
    basicstyle=\ttfamily\footnotesize,
    breakatwhitespace=false,         
    breaklines=true,                 
    captionpos=b,                    
    keepspaces=true,                 
    numbers=left,                    
    numbersep=5pt,                  
    showspaces=false,                
    showstringspaces=false,
    showtabs=false,                  
    tabsize=2
}

\title{Can One Domain Help Others? A Data-Centric Study on Multi-Domain Reasoning via Reinforcement Learning}

\author[1,2\dag]{Yu Li}
\author[1,2\dag]{Zhuoshi Pan}
\author[1,2\dag]{Honglin Lin}
\author[1,2]{Mengyuan Sun}
\author[1,2]{Conghui He}
\author[1,2\ast]{Lijun Wu}

\affiliation[1]{OpenDataLab}
\affiliation[2]{Shanghai Artificial Intelligence Laboratory}

\definecolor{myLavender}{HTML}{e2d5ba}
\definecolor{myDarkBlue}{HTML}{14bc94}
\definecolor{green1}{HTML}{f3eedd}
\definecolor{purple1}{HTML}{303030}
\definecolor{green2}{HTML}{BFF6BA}
\definecolor{blue2}{HTML}{508AB2}
\definecolor{red1}{HTML}{ff6600}
\definecolor{blue1}{HTML}{0085c7}
\newcommand{\rrr}[1]{\textcolor{red1!70}{#1}}
\newcommand{\bbb}[1]{\textcolor{blue1}{#1}}
\newtcolorbox{findbox}[2][]{
    enhanced,
    colback=green2!18,
    colframe=myDarkBlue!50,
    arc=3pt,
    title=#2,
    fonttitle=\bfseries\large,
    coltitle=blue,
    colbacktitle=myDarkBlue!60,
    attach boxed title to top center={yshift=-2mm},
    boxed title style={
        colframe=myDarkBlue,
        arc=3pt,
    },
    coltext=black!85,
    fontupper=\linespread{1.2}\selectfont,
    #1 
}

\newcommand{\myfind}[3][]{%
    \begin{findbox}[#1]{#2}
    #3
    \end{findbox}%
}

\newtcbtheorem[number within=section]{exmp}{Example}%
{beforeafter skip balanced=.01in,colback=green1!30,colframe=purple1!70,fonttitle=\small \bfseries, left=.03in, right=.03in,bottom=.02in, top=.02in}{exmp}

\abstract{
Reinforcement Learning with Verifiable Rewards (RLVR) has emerged as a powerful paradigm for enhancing the reasoning capabilities of large language models (LLMs). 
Existing research has predominantly concentrated on isolated reasoning domains—such as mathematical problem-solving, coding tasks, or logical reasoning.
However, real-world reasoning scenarios inherently demand an integrated application of multiple cognitive skills. Despite this, the interplay among these reasoning skills under reinforcement learning (RL) remains poorly understood. 
To bridge this gap, we present a systematic investigation of multi-domain reasoning within the RLVR framework, explicitly focusing on three primary domains: mathematical reasoning, code generation, and logical puzzle solving. 
We conduct a comprehensive study comprising four key components:
(1) Leveraging the GRPO algorithm and the Qwen-2.5-7B model family, our study thoroughly evaluates the models' in-domain improvements and cross-domain generalization capabilities when trained on single-domain datasets. 
(2) Additionally, we examine the intricate interactions—including mutual enhancements and conflicts—that emerge during combined cross-domain training. 
(3) To further understand the influence of supervised fine-tuning (SFT) on RL, we also analyze and compare performance differences between base and instruct models under identical RL configurations. 
(4) Furthermore, we delve into critical RL training details, systematically exploring the impacts of curriculum learning strategies, variations in reward design, and language-specific factors (e.g., Chinese vs. English datasets). 
Through extensive experiments, our results offer significant insights into the dynamics governing domain interactions, revealing key factors influencing both specialized and generalizable reasoning performance. These findings provide valuable guidance for optimizing RL methodologies to foster comprehensive, multi-domain reasoning capabilities in LLMs.
}

\date{\today}
\correspondence{Lijun Wu, \email{wulijun@pjlab.org.cn}}

\metadata[Code]{\url{https://github.com/Leey21/A-Data-Centric-Study}}
\metadata[Equal contribution]{Yu Li, Zhuoshi Pan, Honglin Lin}

\begin{document}

\maketitle

\section{Introduction}
Recent advances in Reinforcement Learning with Verifiable Rewards (RLVR)~\citep{yu2025rlpr,liu2025prorl,wu2025rlvr,liu2025understanding,yue2025does}, exemplified by DeepSeek-R1-Zero~\citep{guo2025deepseek}, have demonstrated that Reinforcement Learning (RL) can substantially enhance the reasoning capabilities of Large Language Models (LLMs) even without relying on supervised fine-tuning (SFT)~\citep{zhou2025r1,chu2025sft,chen2025sft}. This approach has revealed emergent reasoning capacities, notably through length-dependent performance improvements. Later, multiple studies building upon this framework have validated the effectiveness of RLVR across specialized reasoning domains. For instance, Logic-RL~\citep{xie2025logic} has significantly advanced deductive reasoning, while Open-Reasoner-Zero~\citep{hu2025open} has set new performance benchmarks in mathematical reasoning tasks via RL-driven methods. These successes highlight the broad versatility and effectiveness of RLVR as a post-training framework to enhance reasoning skills across diverse domains.

Despite these breakthroughs, existing research has largely concentrated on reasoning tasks within isolated domains, such as mathematical problem-solving~\citep{zeng2025simplerl}, code generation~\citep{code-r1}, or logical reasoning~\citep{tinyzero} tasks individually. In practice, however, comprehensive reasoning~\citep{li2025cipherbank,pan2025reststresstestinglarge} often demands the seamless integration of multiple cognitive skills. Crucially, the interactions among these reasoning skills under RLVR—particularly regarding how domain-specific training influences cross-domain generalization, training dynamics, reward structures, curriculum strategies, and training languages—have remained underexplored. A comprehensive, systematic investigation of these multi-domain interactions is thus essential to understand and optimize RLVR for holistic reasoning applications.

In this paper, we conduct a systematic study of multi-domain reasoning under the RLVR paradigm, explicitly focusing on three critical reasoning domains: \textit{\textbf{Math}}, \textit{\textbf{Code}}, and \textit{\textbf{Puzzle}}. Leveraging the Group Relative Policy Optimization (GRPO)~\citep{shao2024deepseekmath} algorithm and the Qwen-2.5~\citep{qwen2025qwen25technicalreport} model family, (1) we first examine the impacts of single-domain training on in-domain performance and cross-domain generalization. (2) Then, we identify complex interactions, including mutual enhancements and conflicts, that emerge when integrating multiple domains during training. (3) To further elucidate the role of supervised fine-tuning in enhancing RL effectiveness, we systematically analyze performance differences between base and instruct models. (4) Moreover, our analysis explores critical training strategies, such as curriculum learning, variations in reward design, and language-specific effects (e.g., Chinese versus English training datasets). 

Through rigorous experimental evaluation, we uncover nuanced insights into domain interactions, revealing fundamental mechanisms that influence both domain-specific expertise and generalized reasoning capabilities. Our contributions provide valuable guidelines for future research aiming to refine RL methodologies, ultimately fostering more robust and integrated multi-domain reasoning in LLMs. The primary findings of this study are summarized as follows:

\myfind{Overall Takeaways}{
\begin{itemize}[leftmargin=*, label=\scriptsize$\bullet$, nolistsep]
    \item \textbf{Puzzle and math data provide mutual support.} Logical reasoning and mathematical capabilities complement each other and enhance overall model performance.
    
    \item \textbf{Code reasoning has mixed cross-domain effects.} It strengthens reasoning transfer for the instruct model but may constrain the base model’s reasoning capacity.
    
    \item  \textbf{Cross-domain data leads to more robust performance.} Combining diverse data often results in stronger or more balanced model capabilities, but requires more sophisticated design to address conflicts that may arise between different domains.
    
    \item  \textbf{SFT boosts the effectiveness of RL.} Incorporating an SFT stage before RL leads to substantial improvements in model performance.
    
    \item \textbf{Template consistency is critical.} Misalignment between training and evaluation templates can significantly degrade performance, which also indicates that the robustness of RLVR’s generalization ability is challenged when trained on specific domains.

    \item \textbf{Policy refresh Benefits.} Periodic updates to the reference model and optimizer state in curriculum learning can somewhat improve model stability and performance.
    
    \item \textbf{Reward design should adapt to difficulty.} Tailoring reward settings to how the model performs on the training data can improve learning efficiency.
    
    \item \textbf{RLVR is language-sensitive.} Models trained in Chinese underperforms that trained in English with a consistent performance gap.
\end{itemize}
}

Besides the above overall takeaways, more detailed and throughful observations are illustrated in specific sections in the following studies. 

\section {Experimental Configuration}
This study aims to explore the model's fine-grained reasoning capabilities from a data-centric perspective, through various training approaches including single-domain data training, cross-domain data combination, curriculum learning, different reward settings, and training languages.

\subsection{Multi-Domain Training Setup}

We categorize our reasoning domains into \textit{\textbf{Math}}, \textit{\textbf{Code}}, and \textit{\textbf{Puzzle}}. \textbf{To support multi-domain training}, we curate \textbf{domain-specific datasets} for these areas, as detailed in Table~\ref{tab:datasets}. (1) For the Math domain, we select the popular DeepScaleR (DSR)~\citep{DeepScaleR2025} and CountDown (CD)~\citep{tinyzero}. (2) In the Code domain, our experimental data consists of CodeR1-12k~\citep{code-r1}, which includes 2K reliable LeetCode~\citep{xia2025leetcodedatasettemporaldatasetrobust} data and 10K verified data filtered from 26K TACO~\citep{li2023taco} data. (3) For the Puzzle domain, we focus on two main categories: Knights-and-Knaves (KK)~\citep{xie2024memorization} and Logic Puzzle Baron (LPB)~\citep{logicpuzzlebaron}. Since the LPB dataset lacks ground-truth answers, we utilize DeepSeek-R1~\cite{guo2025deepseek} to annotate 2.4K easy-level puzzles, treating these annotations as pseudo ground truth answer for our RL training. For consistency across domains, we randomly sample larger datasets like DSR (40.3k) and CD (490k) to 10k samples, equalizing the data scale for subsequent training.

\begin{table}[h]
\centering
\caption{Datasets for multi-domain training.}
\label{tab:datasets}
\resizebox{0.75\textwidth}{!}{%
\begin{tabular}{llcc}
\toprule
\textbf{Domain} & \textbf{Training Dataset} & \textbf{Data Size} & \textbf{Reward Scheme} \\
\midrule
\multirow{2}{*}{Math} & DeepScaleR (DSR)~\citep{DeepScaleR2025} & 10k & Binary 0-1  \\
                              & CountDown (CD)~\citep{tinyzero} & 10k & Binary 0-1 \\
\midrule
Code & CodeR1-12k~\citep{code-r1} & 12k & Binary 0-1\\
\midrule
\multirow{2}{*}{Puzzle} & Knights-and-Knaves (KK)~\citep{xie2024memorization} & 5.4k & Binary 0-1 \\
                        & Logic Puzzle Baron (LPB)~\citep{logicpuzzlebaron} & 2.4k & Proportional 0-1 \\
\bottomrule
\end{tabular}
}
\end{table}
\textbf{For the reward,} we design task-specific schemes based on careful analysis of each dataset and the model’s initial performance. The LPB dataset stands out for its higher difficulty: models often fail to produce correct answers in a single attempt at the start of training. To address this, LPB uses a proportional 0–1 reward based on the fraction of correctly predicted cells, while all other datasets adopt a simpler binary 0–1 reward based solely on final answer correctness, without additional format checks. More details on reward design are provided in Section~\ref{reward}.

\textbf{For model selection}, we adopt the Qwen2.5-7B-Base and Qwen2.5-7B-Instruct models as starting points for training. Notably, as emphasized in~\cite{liu2025understanding}, training templates play a crucial role in both the training and testing phases. Our experiments demonstrate that \textbf{template consistency is crucial}; hence, we standardize the use of the \textit{R1-template} (Table \ref{tab:r0_template})~\citep{guo2025deepseek} during training. In testing, we also adopt the \textit{R1-template} to ensure consistency between training and testing phases. We observe that mismatched templates severely degrade model performance, with a detailed analysis provided in Section \ref{dismatch}.

\begin{table}[h]
    \centering
    \small
        \caption{Template for DeepSeek-R1-Zero. \rrr{prompt} will be replaced with the specific reasoning question.}
    \begin{tabular}{l}
    \toprule
\textbf{R1-template:} A conversation between the User and Assistant. The User asks a question, and the Assistant \\
solves it. The Assistant first thinks about the reasoning process internally and then provides the User with \\
the answer. The reasoning process and the answer are enclosed within \textless think\textgreater\ \textless/think\textgreater\ and \textless answer\textgreater\ \\
\textless/answer\textgreater\ tags, respectively, i.e., \textless think\textgreater\ reasoning process here \textless/think\textgreater\ \textless answer\textgreater\ answer here \\ \textless/answer\textgreater. User: \rrr{prompt}. Assistant:\\
     \bottomrule
    \end{tabular}
    \label{tab:r0_template}
\end{table}

\textbf{For the optimization algorithm}, we adopt Group Relative Policy Optimization (GRPO)~\citep{shao2024deepseekmath} as the core RL algorithm. Compared with Proximal Policy Optimization (PPO)~\citep{schulman2017proximalpolicyoptimizationalgorithms}, GRPO dispenses with the traditional value model. Instead, it evaluates the advantage of different responses by assessing the quality differences among answers within a rollout group. GRPO formally maximizes the following objective:

\begin{equation}
  \mathcal{L}_{\text{GRPO}}(\theta) = \mathbb{E}_{\tau \sim \pi_\theta} \Bigg[
\min\Big(
r_\theta(\tau) A(\tau), \\
\quad \text{clip}(r_\theta(\tau), 1 - \epsilon, 1 + \epsilon) A(\tau)
\Big)
\Bigg],  
\label{equ:grpo}
\end{equation}
where $\tau$ denotes the response sampled from the current policy $\pi_\theta$, and $r_\theta(\tau)=\frac{\pi_\theta(\tau)}{\pi_{\text{old}}(\tau)}$ represents the probability ratio between the current policy and the previous policy before each actor update. Unlike PPO, the advantage in GRPO does not depend on a critic model. Instead, it estimates the advantage by calculating a baseline directly from the rollout group’s scores $\{R_i\}_{i\in G(\tau)}$:
\begin{equation}
    A(\tau) = \frac{R_\tau - mean(\{R_i\}_{i\in G(\tau)})}{std(\{R_i\}_{i\in G(\tau)})}.
\end{equation}

\textbf{For Experimental Framework}, all experiments are conducted using the \textbf{veRL} framework~\citep{sheng2024hybridflow}. Both training and testing are conducted on a cluster equipped with 8 × A100 GPUs.

\subsection{Evaluation Settings}

To ensure a comprehensive evaluation of model performance, we employ representative benchmarks across three key domains: mathematical reasoning, code generation, and logical problem-solving.

\textbf{Math Domain:} We evaluate in-domain mathematical reasoning using MATH500~\citep{hendrycksmath2021}, AIME24\footnote{\url{https://huggingface.co/datasets/AI-MO/aimo-validation-aime}}, and CountDown~\citep{tinyzero}. Notably, our implementation of MATH500 adopts a strict 0-shot evaluation, without providing any prior examples—this contrasts with many existing works where the number of shots is often unspecified. For CountDown, we follow the dataset split defined in TinyZero~\citep{tinyzero} and augment it with 24-game dataset (1.36k)\footnote{\url{https://huggingface.co/datasets/nlile/24-game}}, which we also evaluate under a 0-shot setting.

\textbf{Code Domain:} We utilize HumanEval~\citep{chen2021codex} and MBPP~\citep{austin2021program} to evaluate code generation proficiency. For MBPP, we employ 3-shot prompting, while HumanEval is evaluated in a 0-shot setting.

\textbf{Puzzle Domain:} We assess performance using test sets derived from KK (the dataset's own test set) and ZebraLogicBench (Zebra)~\citep{lin2025zebralogicscalinglimitsllms}, which provide diverse scenarios for evaluating logical reasoning abilities. These benchmarks are evaluated in a 0-shot configuration.

\textbf{Evaluation Details:} All evaluations are performed using OpenCompass~\citep{2023opencompass} toolkit, conducted with consistent hyperparameters: temperature = 0.7, top-p = 0.95, and a maximum output length of 8,192 tokens. We emphasize that many existing studies report inconsistent results when reproducing baseline models or conducting new evaluations, often due to misaligned templates or unspecified few-shot configurations. To promote reproducibility, we provide complete details of our prompt templates and few-shot examples in Appendix~\ref{prompt}, and encourage future work to maintain similar transparency in benchmark reporting.

\section{Performance with Single-Domain Data}
In this section, we evaluate the performance of models trained via RL using single-domain data. Our goal is to investigate the impact of single-domain training on both in-domain and out-of-domain (OOD) benchmark performance, providing insights into the models' generalization capabilities. Additionally, findings from single-domain training will guide the design of subsequent experiments involving combined-domain data.
In all experimental results, \bbb{Blue} denotes positive improvements, while \rrr{Orange} indicates a decline relative to the baseline model.

For clarity in the subsequent analysis, we adopt the notation \textit{Base-DSR} and \textit{Instruct-DSR} to represent models trained on the DSR dataset using the Base model and the Instruct model, respectively. The same naming convention applies to other datasets.

\subsection{Math Domain}

\label{math_Domain}

Next, we focus on the math domain, conducting all experiments under identical settings to ensure a fair comparison. Key training hyperparameters are detailed in Table~\ref{tab:hyperparams}. Given that mathematical tasks often require longer Chains of Thought (CoT) for reasoning~\citep{xu2025redstar,wen2025light,pei2025mathfusion,pan2025lemma}, we set a larger max token for training. For brevity, Batch Size is abbreviated as BS.

\begin{table}[h]
\centering
\caption{Key hyperparameters for math domain.}
\label{tab:hyperparams}
\resizebox{0.75\textwidth}{!}{%
\begin{tabular}{*{6}{c}}
\toprule
\textbf{Max Token} & \textbf{Rollout BS} & \textbf{Mini BS} & \textbf{LR} & \textbf{Rollout Times} & \textbf{Epochs} \\
\midrule
8,192 & 256 & 128 & $1 \times 10^{-6}$ & 8 & 12 \\
\bottomrule
\end{tabular}
}
\end{table}

As shown in Table~\ref{tab:math_single}, our results highlight two key findings:

(1) \textbf{RLVR enhances in-domain performance.} Across all math domain experiments, RLVR consistently improves average model performance. For instance, the \textit{Base-DSR} model increases MATH500 accuracy by 19.60 over the base model's 56.40, while the \textit{Base-CD} model boosts CountDown accuracy by 75.56, far surpassing the base model's 1.05. Similar improvements are observed with instruct models. However, both the base and \textit{Base-DSR} models perform poorly on the CountDown dataset, achieving only 1.05 and 0.04, respectively. Analysis of model outputs reveals that the base model struggles to meet the task requirement of using all numbers exactly once, highlighting its limited instruction-following capabilities without specialized training.

(2) \textbf{Math training improves puzzle performance but impairs coding skills. } Math training improves puzzle-solving performance, with \textit{Base-DSR} and \textit{Base-CD} models increasing puzzle averages to 24.08 and 21.13, respectively, from 9.07 (base). However, coding performance declines significantly; for instance, the \textit{Base-CD} model's code performance drops to 29.59 from 67.46. This suggests that while math training enhances puzzle-solving capabilities, it may hinder coding skills due to differing reasoning requirements.

\begin{table}[h]
\centering
\caption{Model performance (\%) after training in the math domain.}
\label{tab:results}
\resizebox{\textwidth}{!}{%
\begin{tabular}{lcccc ccc ccc}
\toprule
\multirow{2}{*}{\textbf{Data}} & \multicolumn{4}{c}{\textbf{Math}} & \multicolumn{3}{c}{\textbf{Code}} & \multicolumn{3}{c}{\textbf{Puzzle}} \\
\cmidrule(r){2-5} \cmidrule(r){6-8} \cmidrule(r){9-11}
 & MATH500 & CountDown & AIME24 & Avg. & HumanEval & MBPP & Avg. & KK & Zebra & Avg. \\

\midrule

Base & 56.40 & 1.05 & 10.00 & 22.48 & 70.12 & \textbf{64.80} & \textbf{67.46} & 17.86 & 0.27 & 9.07 \\
\addlinespace[2pt]
\hdashline
\addlinespace[2pt]
DSR & \textbf{76.00} & 0.04 & 13.33 & \bbb{29.79} & \textbf{82.00} & 39.20 & \rrr{60.60} & \textbf{26.71} & \textbf{21.46} & \textbf{\bbb{24.08}} \\
CD & 67.20 & \textbf{76.61} & 13.33 & \textbf{\bbb{52.38}} & 23.78 & 35.40 & \rrr{29.59} & 21.40 & 21.11 & \bbb{21.13} \\
DSR\&CD & 72.00 & 53.77 & \textbf{16.67} & \bbb{47.48} & 66.46 & 62.00 & \rrr{64.23} & 26.43 & 18.42 & \bbb{22.42} \\
\midrule

Instruct & 69.00 & 24.35 & 13.33 & 35.56 & \bf 82.93 & \bf 62.80 & \bf 72.87 & 10.14 & \bf 31.50 & 20.82 \\
\addlinespace[2pt]
\hdashline
\addlinespace[2pt]
DSR & 72.60 & 35.33 & 10.00 & \bbb{39.31} & 78.05 & 53.60 & \rrr{65.82} & 25.00 & 29.66 & \bf \bbb{27.33} \\
CD & 72.40 & \bf 66.89 & \bf 20.00 & \bf \bbb{53.10} & 79.88 & 61.20 & \rrr{70.54} & 24.29 & 29.36 & \bbb{26.82} \\
DSR\&CD  & \bf 74.60 & 64.79 & 13.33 & \bbb{50.91} & 80.49 & 54.40 & \rrr{67.44} & \bf 26.86 & 24.64 & \bbb{25.75} \\
\bottomrule
\end{tabular}
}
\label{tab:math_single}
\end{table}

\myfind{Takeaway for Math RL}{
\begin{itemize}[leftmargin=*, label=$\bullet$, itemsep=0pt, topsep=0pt, partopsep=0pt]
    \item \textbf{Math training boosts mathematical performance.} Math-focused training significantly enhances model performance on mathematical tasks.
    \item \textbf{Math skills aid puzzles but hinder coding.} Math training improves puzzle-solving abilities through shared logical reasoning but often reduces coding performance.
\end{itemize}
}

\subsection{Code Domain}

In the code domain, RL typically focuses on generating executable code from user-provided instructions and verifying correctness using predefined test cases. This verification requires a secure sandbox environment to safely run generated code and enforce strict execution time limits to prevent timeouts. The key hyperparameters and sandbox configurations used in our experiments are summarized in Table~\ref{tab:code-hyper}. We highlight several key observations below:

\begin{table}[h]
\centering
\caption{Training hyperparameters for code domain.}
\label{tab:code-hyper}
\resizebox{0.9\textwidth}{!}{%
\begin{tabular}{*{8}{c}}
\toprule
\textbf{Max Token} & \textbf{Rollout BS} & \textbf{Mini BS} &\textbf{LR} & \textbf{Rollout Times} & \textbf{Epochs} & \textbf{Sandbox} & \textbf{Timeout(s)}\\
\midrule
4,096 & 128 & 64 & $1 \times 10^{-6}$ & 5 & 15 & FireJail & 30\\
\bottomrule
\end{tabular}
}
\end{table}

\begin{figure}[htbp]
    \centering
    \begin{minipage}{0.45\textwidth}
        \centering
        \includegraphics[width=\textwidth]{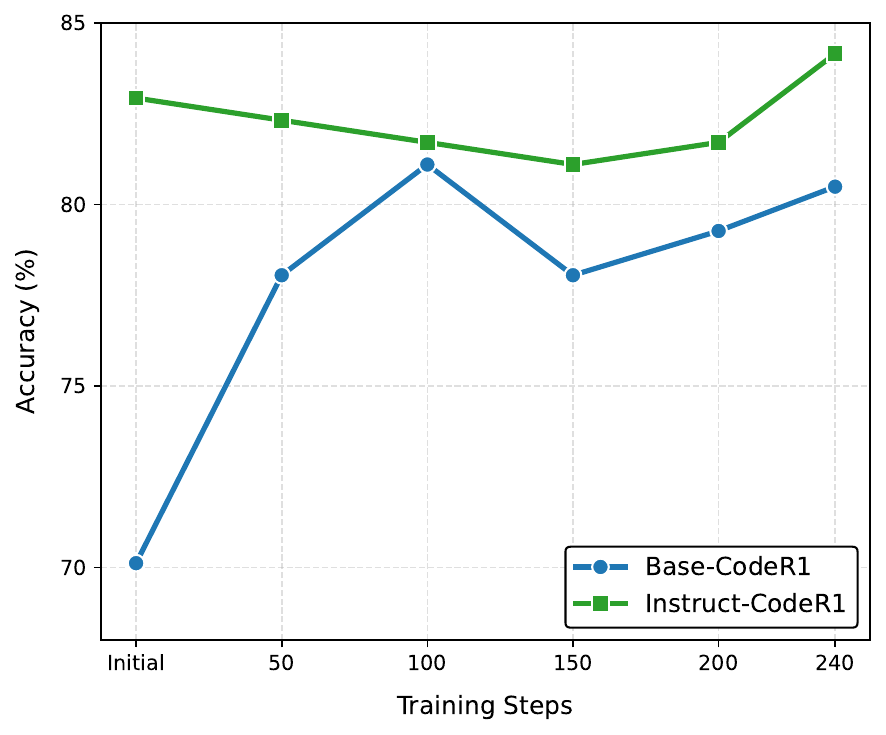}
        \caption{Performance on HumanEval.}
        \label{fig:humaneval}
    \end{minipage}
    \hspace{0.2cm} 
    \begin{minipage}{0.45\textwidth}
        \centering
        \includegraphics[width=\textwidth]{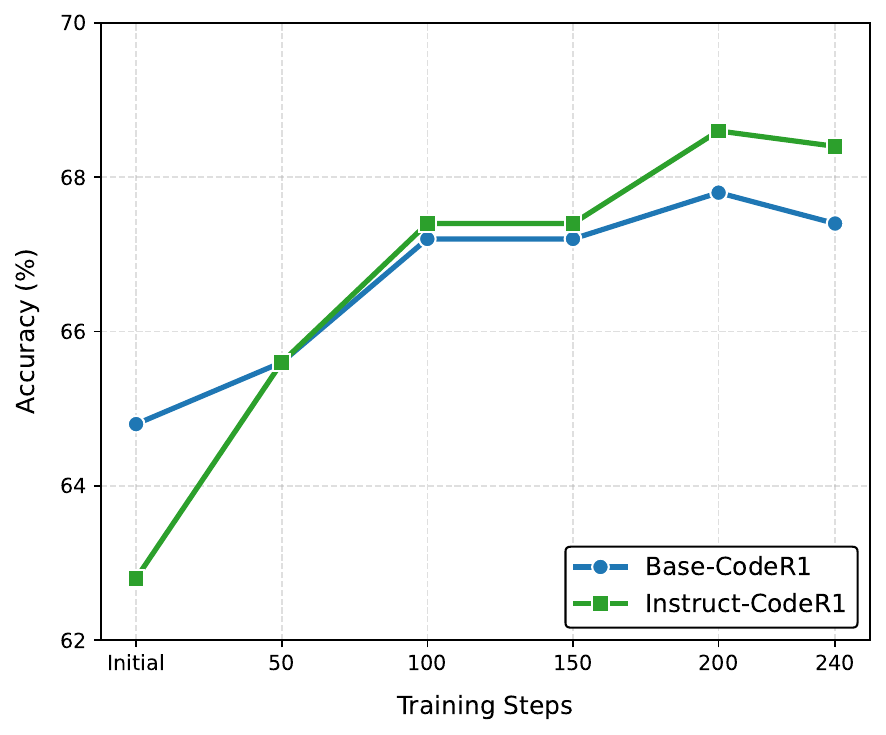}
        \caption{Performance on MBPP.}
        \label{fig:mbpp}
    \end{minipage}
\end{figure}

(1) \textbf{Improvement in in-domain performance.}  
The in-domain performance after RL training on code data is presented in Figures~\ref{fig:humaneval} and~\ref{fig:mbpp}. Both the base and instruct models exhibit substantial improvements on Humaneval and MBPP, demonstrating the efficacy of code data for RL training. The base model, in particular, reveals significant untapped potential, with its Humaneval score surging from 70.12 to 80.49 (+10.37). On MBPP, the base model improves from 64.80 to 67.40. Despite these gains, the instruct model, enhanced by SFT, consistently achieves the highest performance. On Humaneval, the instruct model reaches a superior score of 84.15, improving by 1.83. On MBPP, it overcomes an initial deficit (62.80 vs. 64.80) to attain 68.40, surpassing the base model’s 67.40. This consistent outperformance across both benchmarks underscores the critical importance of SFT in unlocking the full potential of RL training, enabling the instruct model to outperform the base model despite the latter’s remarkable progress.

(2) \textbf{Distinct cross-domain effects of code reasoning.} To examine how enhanced code reasoning influences performance in other domains, we report OOD results in Table~\ref{tab:code-ood}, using checkpoints after 240 training steps. The results reveal contrasting cross-domain effects between the base and instruct models. For the instruct model, improved code reasoning generally brings gains across most OOD benchmarks (except for CountDown). In contrast, the base model shows performance drops on most OOD tasks, except for Zebra. Further analysis of \textit{Base-CodeR1} outputs suggests that the rigid structure of code training data can constrain the base model’s output flexibility, leading to format inconsistencies that hinder correct answer extraction in non-code tasks.

\begin{table}[htbp]
\centering
\caption{Model performance (\%) after training in the code domain.}
\label{tab:code-ood}
\resizebox{0.65\textwidth}{!}{%
\begin{tabular}{l c c c c c}
\toprule
\multirow{2}{*}{Data} & \multicolumn{3}{c}{Math} & \multicolumn{2}{c}{Puzzle} \\
\cmidrule(r){2-4} \cmidrule(r){5-6}
                       & MATH500 & CountDown & AIME24 & KK & Zebra \\
\midrule
Base                  & 56.40   & 1.05      & 10.00  & 17.86 & 0.27 \\
CodeR1  & \rrr{50.80}   & \rrr{0.04}      & \rrr{6.67}   & \rrr{13.85} & \bbb{31.24} \\
\midrule 
Instruct              & 69.00   & 24.35     & 13.33  & 10.14 & 31.50 \\
CodeR1 & \bbb{72.00}   & \rrr{22.59}     & \bbb{16.67}  & \bbb{17.57} & \bbb{32.14} \\
\bottomrule
\end{tabular}
}
\end{table}

\myfind{Takeaway for Code RL}{
\begin{itemize}[leftmargin=*, label=$\bullet$, itemsep=0pt, topsep=0pt, partopsep=0pt]
    \item \textbf{Coding ability enhancement.} Code RL effectively improves the model's ability to handle coding tasks.
\item \textbf{Code reasoning has mixed cross-domain effects.} It strengthens reasoning transfer for the instruct model but may constrain the base model’s reasoning capacity.
\end{itemize}
}

\subsection{Puzzle Domain}
Logic reasoning puzzles require complex logical deduction and multi-step reasoning, posing unique challenges for RL models due to their need for sequential decision-making and pattern recognition. We evaluate RL performance in the puzzle domain using base and instruct models under identical training conditions for fair comparison. Key training hyperparameters are summarized in Table~\ref{tab:puzzle-hyperparams}. 

\begin{table}[h]
\centering
\caption{Key hyperparameters for puzzle domain.}
\label{tab:puzzle-hyperparams}
\resizebox{0.75\textwidth}{!}{%
\begin{tabular}{*{6}{c}}
\toprule
\textbf{Max Token} & \textbf{Rollout BS} & \textbf{Mini BS} &\textbf{LR} & \textbf{Rollout Times} & \textbf{Epochs} \\
\midrule
4,096 & 128& 64 & $1 \times 10^{-6}$ & 5 & 25 \\
\bottomrule
\end{tabular}
}
\end{table}

The results in Table~\ref{tab:puzzle} reveal the following key findings: 

(1) \textbf{Puzzle task performance is substantially enhanced.}
Training on puzzle-specific datasets—namely KK and LPB—significantly boosts performance within the puzzle domain. Exclusive KK training achieves outstanding KK accuracy (94.29 for the base model, 99.14 for the instruct model), while LPB training notably raises Zebra scores (34.60 for the base model, 36.20 for the instruct model). Combining KK and LPB produces more balanced but slightly lower peak performance, with average puzzle accuracies of 61.98 (base) and 59.96 (instruct), indicating that mixed-dataset training offers limited gains beyond single-source specialization.

(2) \textbf{Cross-Domain Generalization from Puzzle Reasoning to Math Tasks.}  
Training on puzzle datasets often enhances the mathematical reasoning ability of models, demonstrating effective transfer of logical skills across domains. For example, training on the KK dataset boosts the base model’s scores to 68.40 on MATH500 and 20.00 on AIME24, approaching the instruct model’s original scores of 69.00 and 13.33, respectively, indicating strong cross-domain generalization. In contrast, the \textit{Instruct-LPB} model exhibits a sharp performance drop on Countdown, falling from 24.35 to 2.47. This decline may stem from LPB’s relatively fixed problem format, which imposes significant constraints on the problem-solving process for Countdown. Tracking this drop, as shown in Table~\ref{single_zebra_inst}, reveals a trend of initial improvement followed by a decline, suggesting that fixed data formats during training can lead to overfitting, thereby limiting the model’s out-of-domain performance.

(3) \textbf{Limited code domain impact.}  
Puzzle training has an inconsistent effect on coding performance. Training on individual datasets often leads to reduced coding scores, but combining KK\&LPB helps mitigate this decline, yielding Code averages of 71.35 (base) and 71.25 (instruct). This is likely due to the mismatch between the fixed format of puzzle data and the requirements of coding tasks. However, the increased \textbf{data diversity} from combining both datasets helps reduce the performance drop seen with single datasets.

\begin{table}[h]
\centering
\caption{Model performance (\%) after training in the puzzle domain.}
\label{tab:puzzle}
\resizebox{\textwidth}{!}{%
\begin{tabular}{lcccccccccc}
\toprule
\multirow{2}{*}{\textbf{Data}}  & \multicolumn{4}{c}{\textbf{Math}} & \multicolumn{3}{c}{\textbf{Code}} & \multicolumn{3}{c}{\textbf{Puzzle}} \\
\cmidrule(r){2-5} \cmidrule(r){6-8} \cmidrule(r){9-11}
& MATH500 & CountDown & AIME24 & Avg. & HumanEval & MBPP & Avg. & KK & Zebra & Avg. \\
\midrule
Base & 56.40 & 1.05 & 10.00 & 22.48 & 70.12 & \textbf{64.80} & 67.46 & 17.86 & 0.27 & 9.07 \\
\addlinespace[2pt]
\hdashline
\addlinespace[2pt]
KK & 68.40 & \textbf{19.36} &\textbf{ 20.00} &  \textbf{\bbb{35.92}} & 60.37 & 51.80 & \rrr{56.09} & \textbf{94.29} & 30.69 &\bf \bbb{62.49}  \\
LPB &\textbf{69.00} & 7.40 & 10.00 &  \bbb{28.80} & 74.40 & 61.60 &  \bbb{68.00} & 16.60 & 34.60 & \bbb{25.60} \\
KK\&LPB & 67.60 & 10.81 & 10.00 &  \bbb{29.47} & \textbf{78.70} & 64.00 &\textbf{  \bbb{71.35} }& 89.29 & \textbf{34.66 }& \bbb{61.98} \\
\midrule
Instruct & 69.00 & 24.35 & 13.33 & 35.56 &\bf 82.93 & 62.80 & \bf 72.87 & 10.14 & 31.50 & 20.82 \\
\addlinespace[2pt]
\hdashline
\addlinespace[2pt]
KK & \bf 73.20 & \bf 33.95 &\bf 23.33 & \bf \bbb{43.49} & 74.39 & 62.80 & \rrr{62.60} &\bf 99.14 & 17.91 & \bbb{58.53} \\
LPB & 69.40 & 2.47 & 13.33 & \rrr{28.41} & 70.70 & \bf 63.80 & \rrr{67.25} & 14.00 & 36.20 & \bbb{25.10} \\
KK\&LPB & 72.40 & 30.30 &\bf 23.33 &  \bbb{42.01} & 80.49 & 62.00 & \rrr{71.25} & 83.29 &\bf 36.62 & \bf \bbb{59.96} \\
\bottomrule
\end{tabular}
}
\end{table}

\myfind{Takeaway for Puzzle RL}{
\begin{itemize}[leftmargin=*, label=$\bullet$, itemsep=0pt, topsep=0pt, partopsep=0pt]
\item \textbf{Puzzle tasks enhance logical reasoning for math tasks.} Puzzle tasks improve logical reasoning, leading to better performance on mathematical tasks. However, this effect does not extend to coding tasks.
\end{itemize}
}

\section{Performance with Combined-Domain Data}

In this section, we reorganize the experimental results of cross-domain RL and provide a systematic analysis of performance across different domain combinations and their interaction patterns. To facilitate a clear examination of how various combinations influence the model's generalization capability and domain-specific task performance, we divide the analysis into two subsections: dual-domain combinations and triple-domain combinations.

Given that many prior studies~\citep{hu2025open, yeo2025demystifying, zeng2025simplerl, tinyzero, xie2025logic} primarily focus on base models, we employ the Qwen2.5-7B Base as the foundation for continued training. For experimental hyperparameters, whenever mathematical data is included, we adopt the same configurations as those used for math tasks (Table~\ref{tab:hyperparams}), in order to accommodate the higher token requirements inherent to math-specific training.

\subsection{Combinations of Dual Domains}

We first examine the model’s performance under pairwise domain combinations, specifically \textit{Math + Puzzle}, \textit{Puzzle + Code}, and \textit{Math + Code}. The results for these configurations are directly compared against their respective single-domain baselines. From the domain perspective, the outcomes are summarized in Table~\ref{tab:two_domain_comb1}.

\begin{table}[h]
\centering
\caption{Performance (\%) of the RL model with $\Delta$ compared to Base.}
\label{tab:two_domain_comb1}
\resizebox{0.8\textwidth}{!}{
\begin{tabular}{lccccccccc}
\toprule
\multirow{2}{*}{\textbf{Training Data}} 
& \multicolumn{2}{c}{\textbf{Math Avg}} 
& \multicolumn{2}{c}{\textbf{Code Avg}} 
& \multicolumn{2}{c}{\textbf{Puzzle Avg}} 
& \multicolumn{2}{c}{\textbf{All Avg}} \\
\cmidrule(r){2-3} \cmidrule(r){4-5} \cmidrule(r){6-7} \cmidrule(r){8-9}
 & Value & $\Delta$ & Value & $\Delta$ & Value & $\Delta$ & Value & $\Delta$ \\
\midrule
Base           & 22.48 & --      & 67.46 & --      & 9.07  & --      & 31.50 & --     \\
\addlinespace[2pt]
\hdashline
\addlinespace[2pt]
Math           & 47.48 & \bbb{+25.00}  & 64.23 & \rrr{-3.23}   & 22.42 & \bbb{+13.35}  & 45.11 & \bbb{+13.61} \\
Puzzle         & 29.47 & \bbb{+6.99}   & 71.35 & \bbb{+3.89}   & 61.98 & \bbb{+52.91}  & 50.72 & \bbb{+19.22} \\
Code           & 19.17 & \rrr{-3.31}   & 73.95 & \bbb{+6.49}   & 22.55 & \bbb{+13.48}  & 35.78 &  \bbb{+4.28} \\
\addlinespace[2pt]
\hdashline
\addlinespace[2pt]
Math + Puzzle  & 49.72 & \bbb{+27.24}  & 44.90 & \rrr{-22.56}  & 49.78 & \bbb{+40.71}  & 48.36& \bbb{+16.86}\\
Puzzle + Code  & 32.06 & \bbb{+9.58}   & 74.88 & \bbb{+7.42}   & 55.15 & \bbb{+46.08}  & 50.89 & \bbb{+19.39}\\
Math + Code    & 47.22 & \bbb{+24.74}  & 75.06 & \bbb{+7.60}   & 25.34 & \bbb{+16.27}  &48.92 & \bbb{+17.42}\\
\bottomrule
\end{tabular}
}
\end{table}

Key observations are summarized as follows:

(1) \textbf{Joint training with specific domain pairs can lead to clear synergistic benefits.} For example, when training with the combination of \textit{Math + Puzzle}, the model’s performance on \textit{Math} improves to 49.72, surpassing the \textit{Math-only} performance of 47.48. Similarly, for \textit{Code} tasks, both additional \textit{Puzzle} and \textit{Math} data lead to improvements in code-related tasks when compared to \textit{Code-only} training. These findings indicate that joint training can facilitate beneficial transfer of knowledge across domains in certain settings.

(2) \textbf{Adding an extra domain does not always lead to better performance and may introduce new generalization challenges.} For the \textit{Puzzle} task, all configurations involving additional domains perform worse than the \textit{Puzzle-only} setting, suggesting that increased data diversity can hinder the model’s ability to specialize in solving puzzles. This also reflects the high degree of specialization required by the \textit{Puzzle} task. Notably, in the \textit{Math + Puzzle} configuration, the model’s performance on \textit{Code} tasks drops significantly, falling below both the \textit{Math-only} and \textit{Puzzle-only} baselines. This may be due to the unique characteristics of the \textit{Code} task, which differs structurally and linguistically from \textit{Math} and \textit{Puzzle}, making generalization more difficult when training data is dominated by other domains. Among all combinations, only \textit{Puzzle + Code} achieves overall strong performance, with an overall improvement of 19.39. These results highlight that incorporating more domains does not guarantee universal improvements and may sometimes impede the model’s ability to adapt to tasks with distinct forms or representations.

\textbf{In summary}, these results show that dual-domain training can yield non-trivial benefits and better task balance in specific settings, but its effectiveness depends on how domains are combined and how training data is allocated. Careful design choices are necessary to leverage synergy and mitigate potential negative interactions.

\subsection{Combinations of Triple Domains}

Furthermore, we investigate the impact of training with data from all domains (\textit{Math + Code + Puzzle}), comparing this setting to the previous optimal configuration (\textit{Puzzle + Code}) in Figure~\ref{3d}. Additionally, Figure~\ref{3d2} illustrates the trends in overall performance as domain combinations are progressively expanded. The following key takeaways emerge:

\begin{figure}[htbp]
\centering
\begin{minipage}{0.45\textwidth}
\centering
\includegraphics[width=\textwidth]{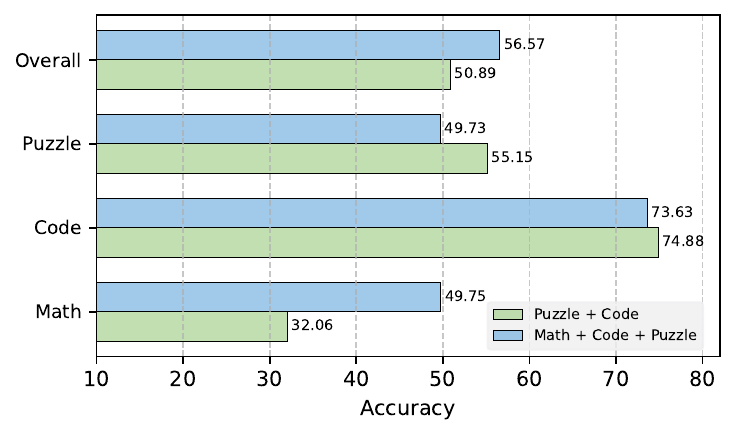}
\caption{Performance comparison of triple-domain and optimal dual-domain data.}
\label{3d}
\end{minipage}
\hspace{0.2cm} 
\begin{minipage}{0.47\textwidth}
\centering
\includegraphics[width=\textwidth]{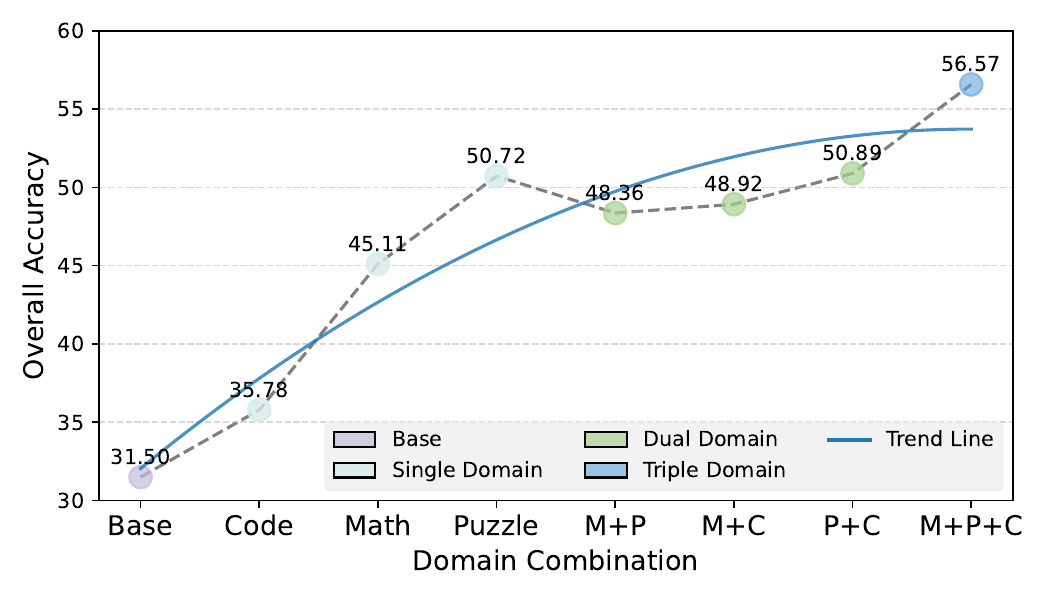}
\caption{Overall model performance under different data combinations.}
\label{3d2}
\end{minipage}
\end{figure}

(1) \textbf{Combining data from all three domains further enhances overall performance, though negative transfer occurs on specific tasks.} As shown in Figure~\ref{3d}, joint training on data from the \textit{Math}, \textit{Code}, and \textit{Puzzle} domains yields an overall average performance of 56.57, surpassing the previous best configuration (\textit{Puzzle + Code}, 50.89). While overall accuracy improves and \textit{Math} tasks reach their highest performance (49.75), the performance on \textit{Puzzle} tasks drops to 49.73, notably lower than the \textit{Puzzle + Code} setting (55.15). This supports our earlier observation that the \textit{Puzzle} domain demands a high degree of specialization, and that incorporating out-of-domain data can negatively impact its results. These results indicate that while broader domain coverage improves overall performance and generalization, the inclusion of \textit{Math} data fails to yield positive effects on \textit{Puzzle} tasks and instead leads to performance degradation. This outcome likely stems from the increased complexity introduced by the additional domain, which adversely impacts specific tasks.

(2) \textbf{Enhanced data diversity contributes to further model performance improvements.} As shown in Figure~\ref{3d2}, excluding the outlier (50.72) observed with the \textit{Puzzle-only} data, the model’s overall performance exhibits a positive trend as domain combinations increase. However, the unusually high value in the \textit{Puzzle} task is primarily due to the model’s exceptional performance on the \textit{KK} task (94.29), which disproportionately elevates the overall score. This outlier does not necessarily reflect balanced performance improvements across all tasks, but rather is driven by a particularly strong result in one specific sub-task.

(3) \textbf{The triple-domain combination improves performance balance across tasks compared to certain dual-domain combinations.} Unlike some dual-domain configurations (e.g., \textit{Math + Puzzle}), which experience significant performance degradation on \textit{Code} tasks (–22.56), the triple-domain approach maintains more balanced performance across all tasks. While some task-specific specialization may be slightly reduced, the inclusion of \textit{Code} data ensures that performance on the \textit{Code} task remains strong and consistent. These results indicate that expanding domain coverage can mitigate the performance collapse on specific tasks, achieving a more stable and generalized performance across tasks.

\textbf{In summary}, the triple-domain combination exemplifies the balanced advantages of multi-skill training. Although incorporating additional domains may lead to a modest reduction in peak performance for individual tasks, this approach achieves the highest overall performance while maintaining competitive results across all subtasks. These findings underscore the substantial potential of carefully designed multi-domain training strategies in building versatile models with broad adaptability.

\myfind{Takeaway for Dual-Domain and Triple-Domain Combinations}{
\begin{itemize}[leftmargin=*, label=$\bullet$, itemsep=0pt, topsep=0pt, partopsep=0pt]
\item \textbf{Multi-domain training improves overall performance.} Combining multiple domains generally leads to better overall performance, with the triple-domain combination showing moderate gains.

\item \textbf{Multi-domain training improves task balance and overall stability.} By providing broader coverage, multi-domain setups help maintain consistent performance across tasks, preventing extreme drops in any single area and promoting more robust, generalized models.

\end{itemize}
}

\section{Evaluating Template Variations in Reinforcement Learning}
\label{dismatch}
A commonly overlooked issue in RL is the mismatch between templates used for training and those applied during testing~\citep{he2024doespromptformattingimpact,wang2025template,jiang2025chatbugcommonvulnerabilityaligned}. Such discrepancies can significantly degrade model performance during evaluation. For instance, training data may utilize an R1-style template, while testing might employ a Qwen template (the default for many Qwen-series models~\citep{yang2024qwen2}) or no template at all, as some evaluation tools, such as OpenCompass~\citep{2023opencompass}, default to a blank template for base model testing (see Table~\ref{tab:template2}).
In this section, we train base and instruct models on the KK dataset using the R1-style template and evaluate their performance with various templates. The results, presented in Table~\ref{tab:template_mismatch_results}, assess the impact of mismatched training and testing templates, highlighting the models' sensitivity to template alignment.
\begin{table}[h]
    \centering
    \small
    \begin{tabular}{l}
    \toprule

\textbf{Qwen Template:} $<$\textbar im\_start\textbar$>$system\textbackslash nPlease reason step by step, and put your final answer within \\
\textbackslash \textbackslash boxed\{\}.$<$\textbar im\_end\textbar$>$\textbackslash n$<$\textbar im\_start \textbar$>$user\textbackslash n\rrr{\{question\}}$<$\textbar im\_end\textbar$>$\textbackslash n$<$\textbar im\_start\textbar$>$ assistant\textbackslash n \\
\textbf{Base Template:} \rrr{\{question\}} \\
     \bottomrule
    \end{tabular}
    \caption{Template for qwen and base model. \rrr{question} will be replaced with the specific reasoning question during training.}
    \label{tab:template2}
    \vspace{-2mm}
\end{table}

\begin{table}[htbp]
  \centering
  \caption{Model performance under different templates.}
  \label{tab:template_mismatch_results}
  
  \resizebox{0.9\textwidth}{!}{%
  \begin{tabular}{llccccccc}
    \toprule
    \multirow{2}{*}{\textbf{Model}} & \multirow{2}{*}{\textbf{Template}} & 
    \multicolumn{3}{c}{\textbf{Math}} & 
    \multicolumn{2}{c}{\textbf{Code}} & 
    \multicolumn{2}{c}{\textbf{Puzzle}} \\
    
    \cmidrule(lr){3-5} \cmidrule(lr){6-7} \cmidrule(lr){8-9}
    & & MATH500 & CountDown & AIME24 & HumanEval & MBPP & KK & Zebra \\
    
    \midrule
    \multirow{3}{*}{\textbf{Base}} & Base Template  & \bbb{72.80} & 0.00   & 6.67   & 60.98 & 3.00 & 31.29 & 16.18 \\
     & Qwen Template  & 69.20 & \bbb{20.79}  & \bbb{13.33}  & \bbb{63.78} & \bbb{64.40} & 94.00 & 0.56  \\
     & R1 Template    & 68.40 & 19.36  & 10.00  & 60.37 & 51.80 & \bbb{94.29} & \bbb{30.69} \\
    
    \midrule
    \multirow{3}{*}{\textbf{Instruct}} & Base Template  & 3.20  & 0.63   & 3.33   & 51.22 & 1.60 & 41.57 & \bbb{21.22} \\
     & Qwen Template  & 1.80  & 0.29   & 0.00   & 32.93 & 42.60 & 62.71 & 9.84  \\
     & R1 Template    & \bbb{73.20} & \bbb{33.95}  & \bbb{23.33}  & \bbb{74.39} & \bbb{62.80} & \bbb{99.14} & 17.91 \\
    \bottomrule
  \end{tabular}%
  }
  \vspace{-1mm}
\end{table}

\begin{wrapfigure}{r}{0.4\linewidth}
    \centering
    \includegraphics[width=\linewidth]{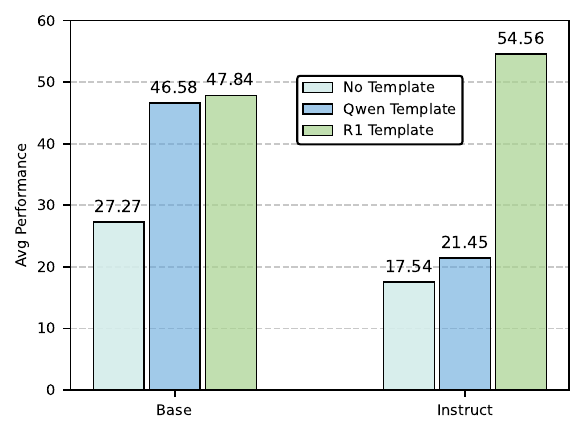}
    \caption{The average test performance of base and instruct models on different templates.}
    \label{fig:template}
    \vspace{-0.5cm}
\end{wrapfigure}
As shown in Table \ref{tab:template_mismatch_results}, a mismatch in templates significantly impacts the model’s performance. We can break this down into the following points:

(1) \textbf{Mismatched Templates Significantly Impact Model Performance.} 
Mismatched templates substantially reduce the performance of base and instruct models across diverse tasks. For the base model, a mismatched base template lowers scores to 0, 3.00, and 31.29 on Countdown, MBPP, and KK, respectively, while a mismatched instruct template decreases the Zebra score to 0.56. Similarly, the instruct model experiences performance drops to 1.80 on MATH500 and 0.29 on Countdown with a mismatched template like Qwen, among other tasks. These declines highlight the models' sensitivity to template mismatches.

(2) \textbf{Matched Templates Typically Achieve Optimal Model Performance.}  
Matched templates consistently enhance the average performance of base and instruct models across benchmarks. As shown in Figure~\ref{fig:template}, the R1 template produces scores of 47.84 and 54.56 for the base and instruct models, respectively, surpassing mismatched conditions. Although no single template excels in every task, the superior performance of matched templates underscores their critical role in ensuring stable and effective outcomes, especially for intricate datasets like KK and Zebra.


\myfind{Takeaway for Template}{
\begin{itemize}[leftmargin=*, label=$\bullet$, itemsep=4pt, topsep=2pt, partopsep=0pt]
    
 \item \textbf{Template consistency is critical.} Mismatched templates degrade model performance on certain tasks, highlighting the current lack of robustness in RLVR.

\end{itemize}
}

\section{The Role of Curriculum Learning in Reinforcement Learning}
\label{curriculum_learning}

While curriculum learning is well-established in SFT~\citep{chen2025self,kong2021adaptive,huang2020curricularface}, its application in RLVR remains insufficiently explored~\citep{parashar2025curriculum,wang2025dump}. To address this gap, we systematically investigate curriculum learning strategies in the Puzzle domain, leveraging the KK dataset. This dataset features clearly defined difficulty variations based on the number of sub-questions per problem, enabling us to effectively categorize data by difficulty levels. This facilitates a focused evaluation of the generalizability of our approach within a specific cognitive challenge.

\textbf{Difficulty stratification and curriculum design:}
Effective curriculum learning hinges on accurately quantifying task difficulty. For the puzzle task, difficulty is defined by the number of sub-questions per problem (PPL), ranging from 3PPL to 8PPL across six levels. Training progresses sequentially from easier to harder tasks, enabling the model to build proficiency in simpler reasoning before tackling more complex problems.

We propose a novel \textbf{policy refresh} strategy, where after each training stage (175 steps), corresponding to a specific difficulty level, the reference model is updated by replacing it with the latest actor model. Additionally, the optimizer state is reset to prevent overfitting to prior difficulty levels. This strategy provides the model with a fresh starting point at each new difficulty level, facilitating stable learning and better adaptation to progressively challenging tasks. The corresponding experimental results are presented in Figure~\ref{fig:cul_kk}. The analysis is as follows:

\begin{figure}[htbp]
    \centering
    \includegraphics[width=1\textwidth]{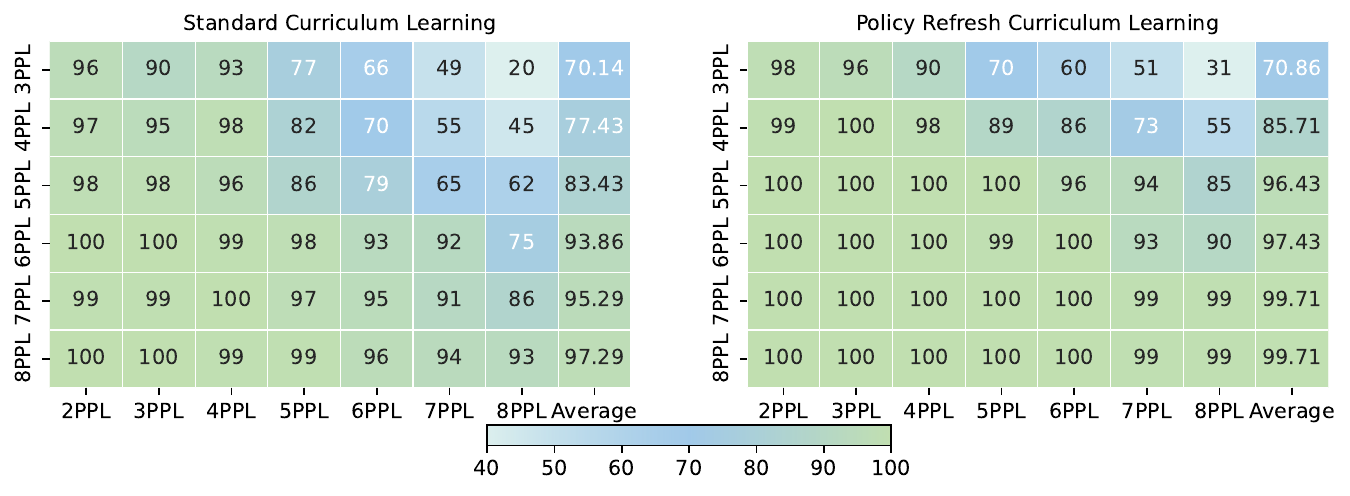}
\caption{Model performance on the KK dataset with different curriculum settings. The x-axis represents the KK difficulty levels, and the y-axis shows the training data sequence from 3PPL to 8PPL.}
    \label{fig:cul_kk}
\end{figure}

(1) \textbf{Curriculum learning improves the upper bound of model performance.} Figure~\ref{fig:cul_kk} demonstrates that curriculum learning under both settings effectively enhances the model’s upper performance bound, achieving accuracies of 97.29 and 99.71, respectively, which significantly surpass the 94.29 accuracy obtained under mixed training. This improvement highlights the key advantages of curriculum learning, including more structured and progressive learning patterns that enable the model to better capture complex task dependencies and enhance generalization capabilities.

(2) \textbf{Policy refresh further improves the performance and convergence rate of curriculum learning.} The right panel of Figure~\ref{fig:cul_kk} reveals that policy refresh accelerates convergence while delivering superior final performance. Beginning from the second stage, models incorporating policy refresh consistently outperform standard curriculum learning, achieving 97.43 accuracy at 6PPL—already exceeding the latter’s final score of 97.29. Remarkably, the policy refresh model nearly reaches perfect accuracy, even surpassing the final result of the instruct model under mixed training (99.14).

\myfind{Takeaway for Curriculum Learning} {
\begin{itemize}[leftmargin=*, label=$\bullet$, itemsep=0pt, topsep=0pt, partopsep=0pt]
    \item \textbf{Curriculum learning demonstrates effectiveness and achieves additional improvements through policy refresh.} Staged training raises the model’s performance upper bound, while periodic reference model updates further accelerate convergence and enhance final results.
\end{itemize}
}

\section{Impact of Reward Styles on Model Performance}
\label{reward}

In this section, we investigate how different reward styles affect model performance using the KK and LPB datasets, selected for their complex problem structures involving multiple interdependent entities. Unlike typical Math and Code datasets, which often feature problems with a single correct answer and rely predominantly on binary reward schemes, the KK and LPB present unique challenges. Each problem requires filling multiple blanks, resembling a cloze task, enabling evaluation of diverse reward strategies. We compare two primary schemes: a binary reward (R1), granting credit only for fully correct responses, and a partial reward (R2), based on the fraction of correctly filled blanks. Additionally, we explore a format reward (R3) using the \texttt{<think>} tag to promote intermediate reasoning, and a rescaled reward (R4) that extends the reward range to $[-1, 1]$ to penalize incorrect responses.

(1) \textbf{Formally, for the KK dataset}, we define the reward function \( R(\text{response}) \) as follows:

\begin{align*}
R(\text{response}) = \left\{
\begin{array}{ll}
\lfloor N_{c}/N\rfloor & \text{(Binary Reward, R1)} \\
N_{c}/N & \text{(Partial Reward, R2)} \\
\lfloor N_{c}/N\rfloor + \text{format reward} & \text{(Format Reward, R3)} \\
1 \text{ if } N_{c} = N, \, -1 \text{ otherwise} & \text{(Rescaled Reward, R4)},
\end{array}
\right.
\label{eq:reward_functions_for_kk}
\end{align*}

where \(N_c\) denotes the number of blanks correctly completed by the model and \(N\) represents the total number of blanks in the puzzle. 

For the LPB dataset, the reward functions differ slightly from those defined for the KK puzzle dataset. Specifically, the forms of the binary reward (R1) and the partial reward (R2) remain identical. However, two modifications are introduced:  (1) the format reward (R3) is modified from $\lfloor N_{c}/N\rfloor + \text{format reward}$ to $N_{c}/N + \text{format reward}$, and the rescaled reward (R4) is defined by linearly scaling the partial reward (R2) to a continuous range from $-1$ to $1$, leading to $2\times\left(N_{c}/N-0.5\right)$.

(2) \textbf{Finally, the reward function for the LPB dataset} is defined as follows:

\begin{align*}
R(\text{response}) = \left\{
\begin{array}{ll}
\lfloor N_{c}/N\rfloor & \text{(Binary Reward, R1)} \\
 N_{c}/N & \text{(Partial Reward, R2)} \\
 N_{c}/N + \text{format reward} & \text{(Format Reward, R3)} \\
2\times(N_{c}/N - 0.5)  & \text{(Rescaled Reward, R4)},
\end{array}
\right.
\end{align*}

The rationale behind these adjustments is supported by empirical results on the LPB  dataset, where the partial reward (R2)—which measures the proportion of correctly filled blanks—yields the best performance, whereas the binary reward (R1) results in the poorest performance. Consequently, we convert the first component of the format reward (R3) from binary to partial, 
and adjust the rescaled reward (R4) into a continuous form to avoid using discrete rewards. To systematically evaluate the impact of different reward styles, we conducted a comprehensive analysis across both in-domain and out-of-domain settings. 

\subsection{Impact of Reward Models on In-Domain Performance}

\begin{figure}[htbp]
    \centering
    \begin{minipage}{0.45\textwidth}
        \centering
        \includegraphics[width=\textwidth]{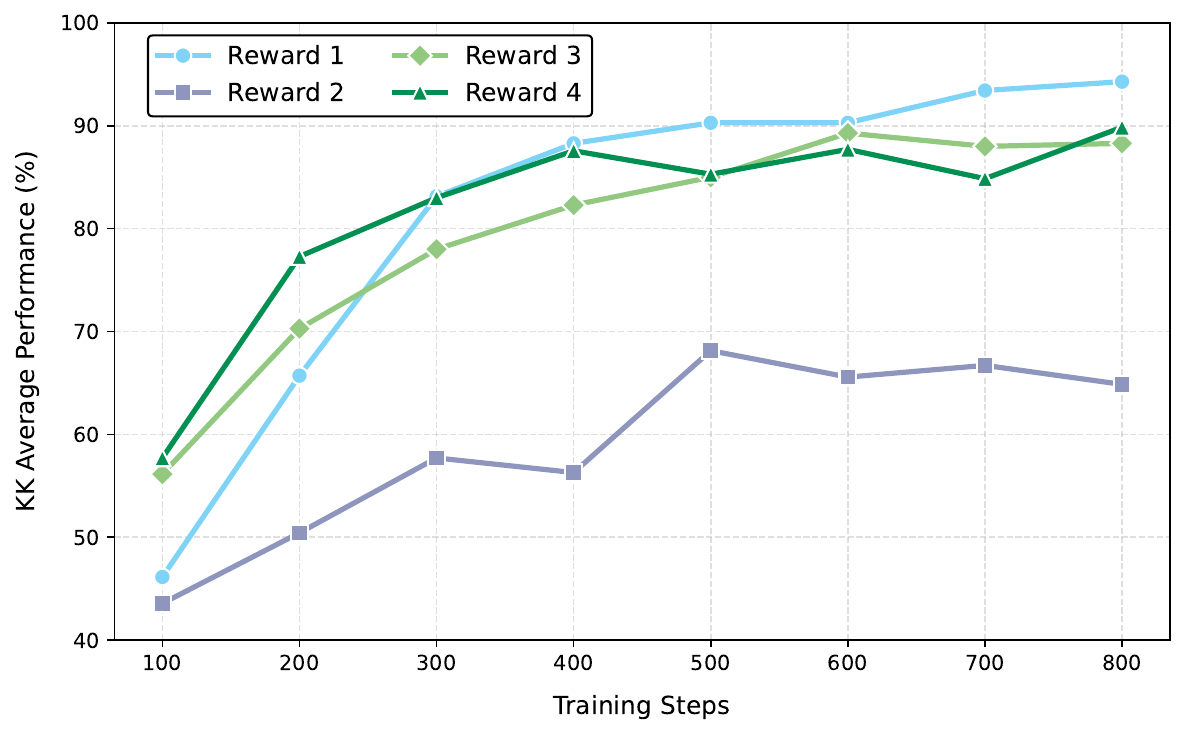}
        \caption{Performance on KK.}
        \label{fig:reward_kk_in}
    \end{minipage}
    \hspace{0.2cm} 
    \begin{minipage}{0.45\textwidth}
        \centering
        \includegraphics[width=\textwidth]{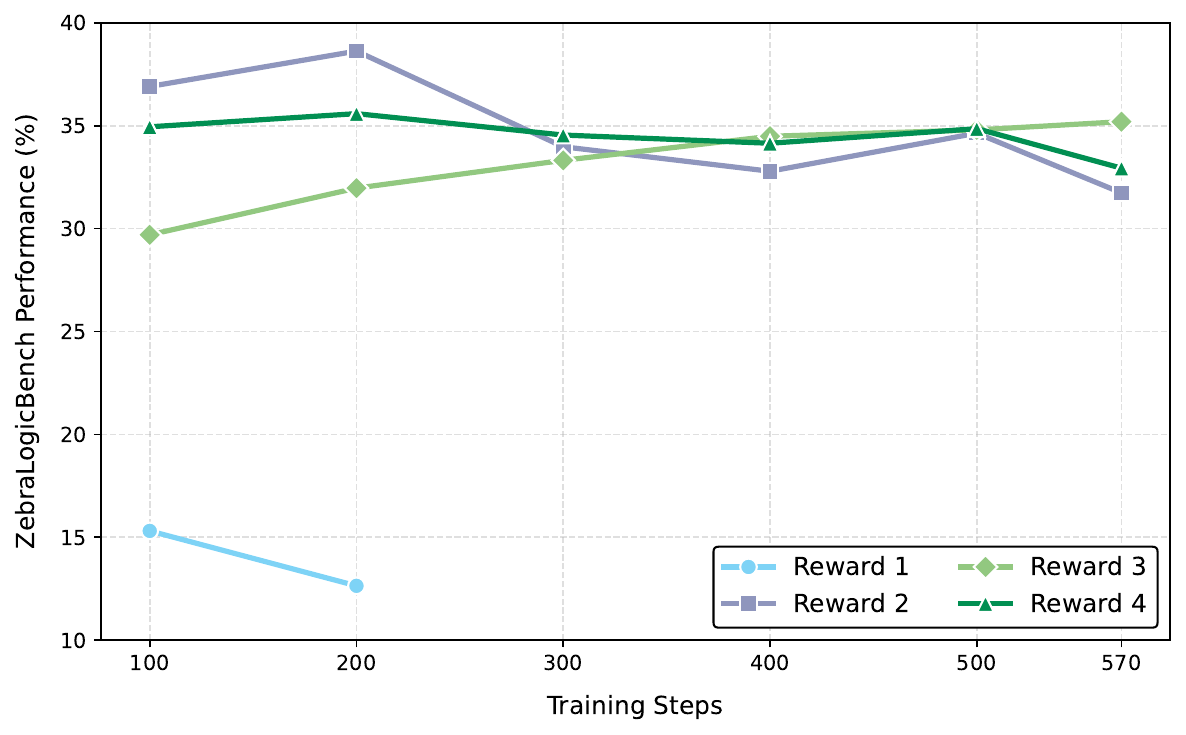}
        \caption{Performance on LPB.}
        \label{fig:reward_zebra_in}
    \end{minipage}
\end{figure}

To investigate the impact of different reward styles on model training, we conduct experiments where only the reward varies, training for a total of 800 steps (or fewer if collapse occurs). We compare performance across the KK and LPB, with results shown in Figures~\ref{fig:reward_kk_in} and \ref{fig:reward_zebra_in}. These comparisons reveal that reward efficacy is highly dataset-dependent, as summarized in the following key findings:

(1) \textbf{Binary reward excels on KK but fails on LPB due to sparsity differences.} On KK, the simplest R1 achieves the best final performance, outperforming more nuanced alternatives by providing clear, direct signals. This aligns with the ``proportional reward trap'' phenomenon~\cite{rastogi2025magistral} in programming tasks, where R2 introduce noise. In  contrast, R1 leads to consistent training collapse on LPB, where extreme reward sparsity arises because the model rarely predicts all puzzle cells correctly, yielding few positive signals. We limit R1 training on LPB to 200 steps to avoid unnecessary computation. This observation underscores that binary rewards are well-suited for relatively easier tasks, like KK, where the base model can achieve complete success sometimes, but are untenable for harder ones like LPB.

(2) \textbf{Partial reward underperforms on KK but offers a viable baseline on LPB, though with limitations.} On KK, R2 shows no advantage over R1 and ultimately degrades performance by injecting noisy learning signals. Conversely, on LPB, R2 emerges as a feasible alternative to R1's collapse, delivering initial promise with a peak accuracy of 38.63 at 200 steps. However, its gains are not sustained, as it fails to accurately penalize the specific erroneous cells in the response, leading to a slight decline. This highlights R2's utility in sparse settings but exposes its inadequacy compared to more sophisticated reward mechanism.

(3) \textbf{Format reward and rescaled reward excel on LPB, while falling short on KK.} On KK, despite early gains from format correction or error suppression, both R3 and R4 yield inferior final performance to R1, indicating that their added complexity does not pay off in domains favoring binary signals. In contrast, on LPB, R3 and R4 initially trail R2 but eventually surpass it, benefiting from more informative signals: R3 stabilizes training via well-formed outputs, and R4 amplifies behavioral differences for better optimization.

Overall, these contrasts demonstrate that optimal reward design is not universal but critically tied to dataset characteristics like sparsity and task complexity. These factors must be carefully considered when designing RLVR training.

\subsection{Impact of Reward Models on OOD Generalization}

\begin{figure}[htbp]
    \centering
    \includegraphics[width=1\textwidth]{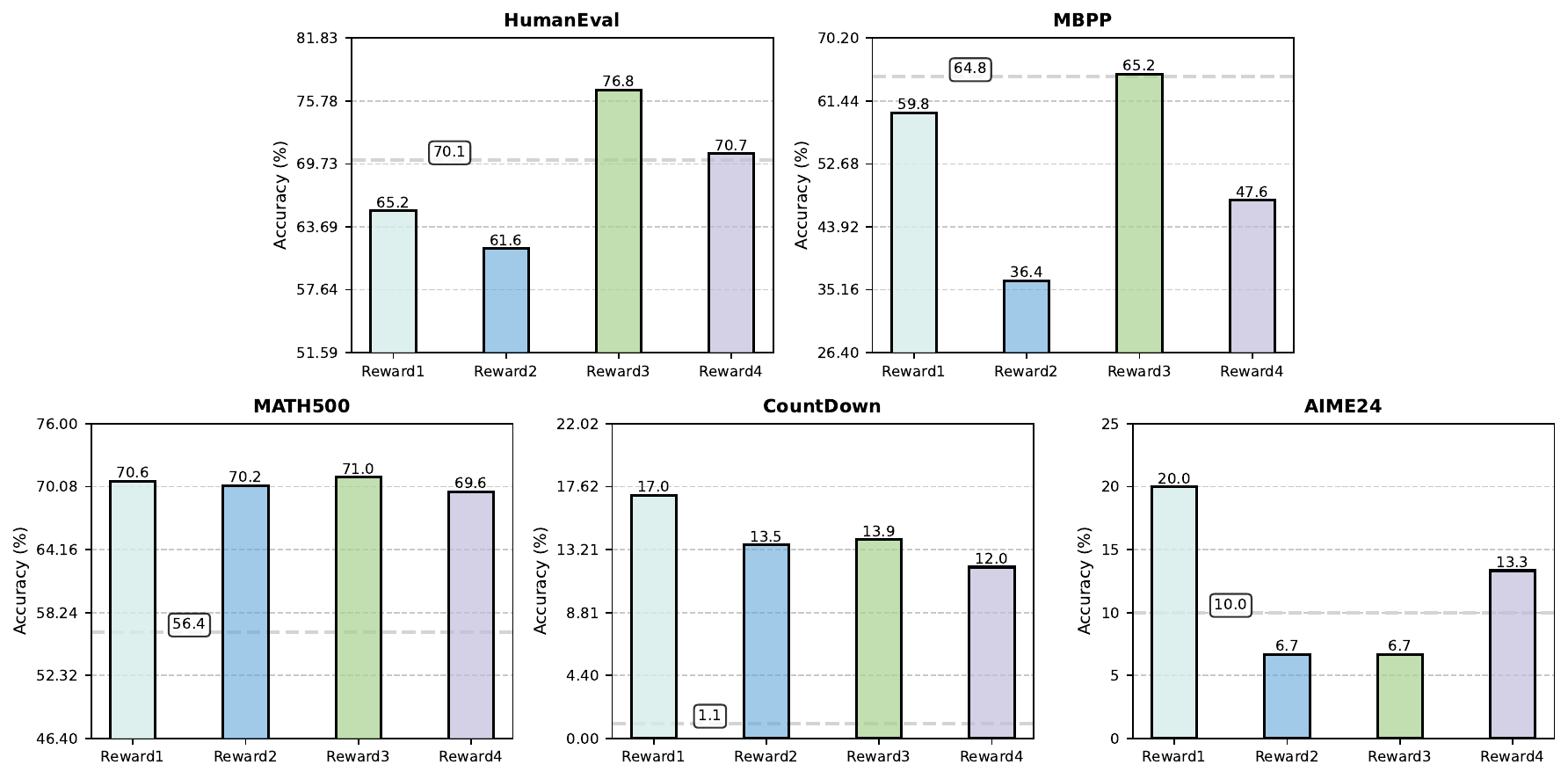}
    \caption{KK-impact of reward configurations (base model shown with dashed lines).}
    \label{fig:reward_kk_ood}
\end{figure}

\begin{figure}[htbp]
    \centering
    \includegraphics[width=1\textwidth]{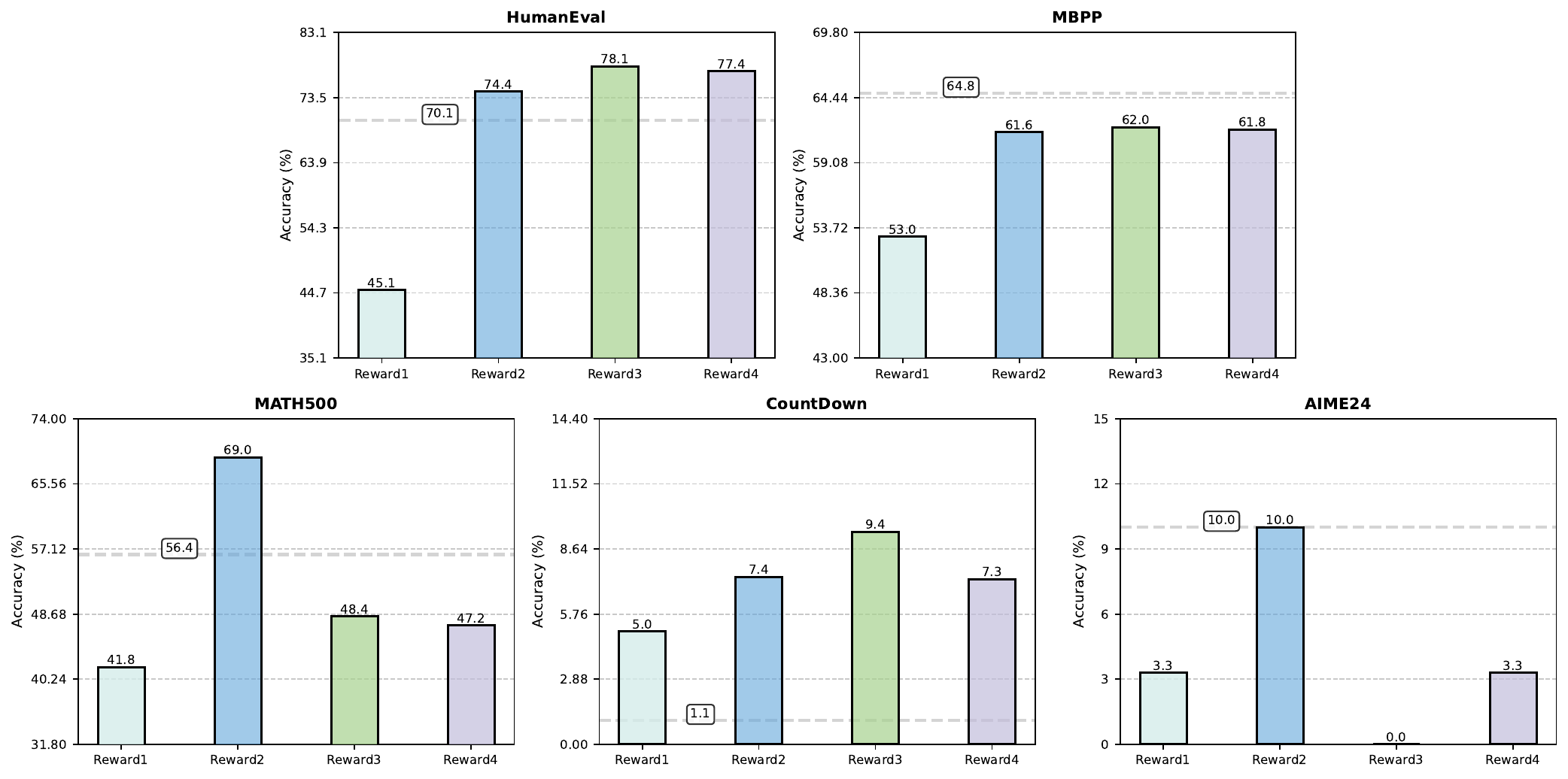}
    \caption{LPB-impact of reward configurations (base model shown with dashed lines).}
    \label{fig:reward_zebra_ood}
\end{figure}

Figures~\ref{fig:reward_kk_ood} and \ref{fig:reward_zebra_ood} illustrate the effects of various reward schemes on OOD tasks, highlighting how reward efficacy varies across datasets. Our key
observations reveal some insightful findings:

(1) \textbf{In mathematical reasoning tasks, reward schemes yield different outcomes depending on the training data.} For KK, all rewards produce similar performance, with the R3 offering no clear advantage over the base model. This contradicts prior claims~\citep{xie2025logic} about the benefits of structured reasoning incentives. In contrast, for LPB, the choice of reward significantly impacts performance: R2 achieves the highest accuracies (e.g., outperforming others on both AIME24 and MATH500 benchmarks), while R1 leads to substantial declines (e.g., MATH500 accuracy dropping from 56.4 to 41.8). 

(2) \textbf{In code generation tasks, different datasets exhibit distinct reward sensitivities.}
For example, training on KK is relatively sensitive to reward design, with the performance of different rewards varying considerably. Most rewards generally degrade performance compared to the base model, though R3 effectively mitigates this negative impact and provides modest improvements. In contrast, training on LPB is less sensitive to reward design: most rewards (excluding R1) perform similarly, yielding gains on the HumanEval benchmark but experiencing drops on the MBPP benchmark. R1, however, suffers from significantly worse performance on both HumanEval and MBPP, which further aligns with the observed in-domain training collapse on LPB.

(3) \textbf{A significant limitation lies in current reward mechanisms.}
These comparisons expose a critical limitation in current reward mechanisms: they operate at the response level rather than the cell level. This means they fail to accurately penalize the erroneous predicted cells but treat all cells equally within a response. This issue is particularly evident in KK, where even R2 led to poor outcomes, aligning with challenges noted in~\cite{rastogi2025magistral}. Overall, the absence of a universal reward strategy emphasizes the task-dependent nature of reward design and the essential role of data diversity in RL training. Optimal rewards should be tailored to specific dataset characteristics—e.g., finer-grained, cell-level schemes for datasets like KK—to overcome the limitations of current response-level rewards.

\myfind{Takeaway for Reward}{
\begin{itemize}[leftmargin=*, label=$\bullet$, itemsep=0pt, topsep=0pt, partopsep=0pt]
    \item \textbf{Reward design should match task complexity.} Binary rewards work well for simpler tasks, while partial rewards are more suitable for complex tasks.
    \item \textbf{Current partial rewards lack precision.} Designing more fine-grained partial reward signals is a promising direction for further improvement.
\end{itemize}
}

\section{Influence of Training Language}

To investigate the impact of training data language~\citep{yu2023largelanguagemodelattributed}, we translate the DeepScaleR dataset into Chinese using \textit{GPT-4.1-nano}\footnote{\url{https://openai.com/index/gpt-4-1/}} and performed RL training with the identical hyperparameters to those used for the English version. 

\begin{wrapfigure}{r}{0.4\linewidth}
    \centering
    \includegraphics[width=\linewidth]{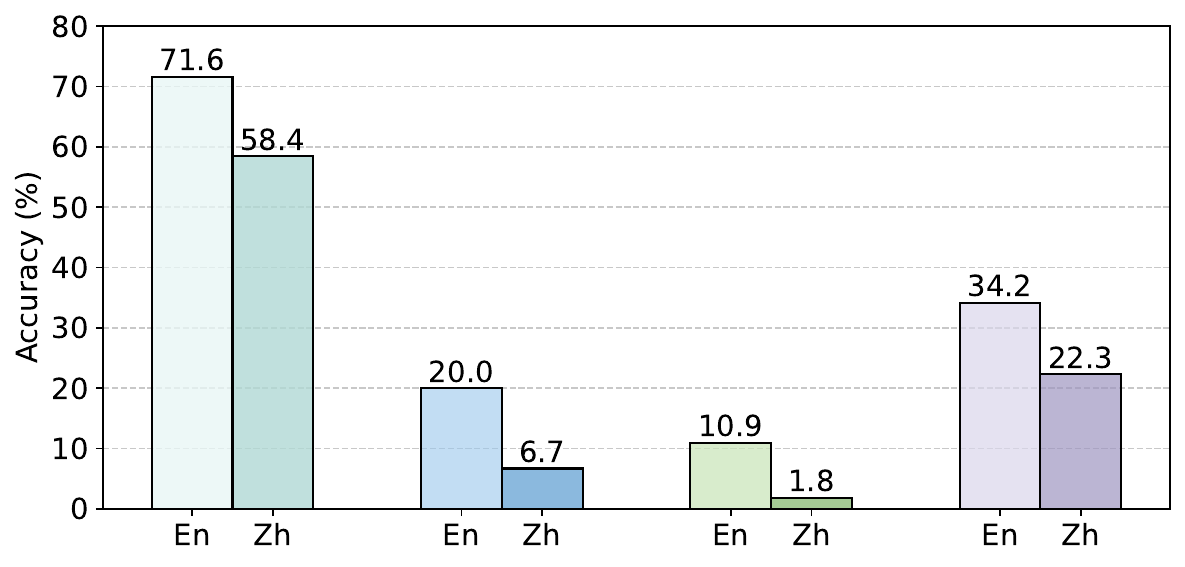}
    \caption{The effect of different training language.}
    \label{fig:language}
    \vspace{-3mm}
\end{wrapfigure}
To ensure the model uses Chinese for reasoning during training, we employ the ``langid''\footnote{\url{https://pypi.org/project/langid/}} package to detect the language of each rollout trajectory.
A reward of 1 is given only when the language is Chinese and the final answer is correct. 
If the language is not Chinese, 
even if the final answer is correct, the reward is 0.
This strict reward function is necessary, as we observe that without it, the model would default to reasoning in English even when the questions are translated into Chinese.
Figure~\ref{fig:language} illustrates the impact of different training languages. It is clear that models trained to reason in Chinese consistently perform worse than their English counterparts. This inefficiency in Chinese language reasoning highlights the need for more advanced post-training algorithms capable of improving cross-lingual generalization for complex reasoning tasks.

\myfind{Takeaway for Training Language}{
\begin{itemize}[leftmargin=*, label=$\bullet$, itemsep=0pt, topsep=0pt, partopsep=0pt]
    \item \textbf{RLVR is language-sensitive.} Models trained in Chinese underperforms that trained in English with a consistent performance gap.
\end{itemize}
}

\section{Conclusion}
In this work, we focus on the reasoning capabilities of large language models, with a data-centric approach. We classify reasoning data into three main domains: Math, Code, and Puzzle. A detailed discussion is provided on the effects of domain-specific data in reinforcement learning and the generalization to out-of-domain data. Through cross-domain data combinations, we reveal the potential interactions between different model capabilities, including both assisting and conflicting phenomena. Additionally, we thoroughly discuss the impact of various factors on model performance in 
Reinforcement Learning, including model template, curriculum learning, reward styles, and training languages, offering some insights from multiple perspectives.

For future work, we believe it would be beneficial to further refine the categorization of reasoning capabilities. For example, introducing data from the science and general reasoning domains would allow for more detailed discussions on data combinations. Furthermore, due to hardware limitations, this paper primarily focuses on the base and instruct models of Qwen 2.5. Future research could expand on the impact of different datasets on models like Llama and DeepSeek. We also believe that RLVR will become a significant milestone in the development of large language models, with data remaining the cornerstone of any training process. We hope that future work will delve deeper into exploring the impact of data on RLVR.


\clearpage
\newpage
\bibliographystyle{plainnat}
\setcitestyle{numbers}
\bibliography{paper}

\clearpage
\newpage
\beginappendix

\section{Step-Level Performance Results for Each Training Dataset}

In the following figures, we present detailed performance results for the base and instruct models across different training datasets. The results show performance at each evaluation checkpoint, illustrating the models’ progression and how different training data influence overall performance.
\begin{itemize}
    \item \textbf{DeepscaleR}: Figures~\ref{single_deepscaler1} and ~\ref{single_deepscaler2}  illustrate the training dynamics on the DeepscaleR dataset. The results show that the model’s mathematical reasoning ability improves consistently and partially generalizes to logical reasoning tasks, while its code reasoning capability drops significantly.

    \begin{figure}[h]
    \centering
    \includegraphics[width=0.95\textwidth]{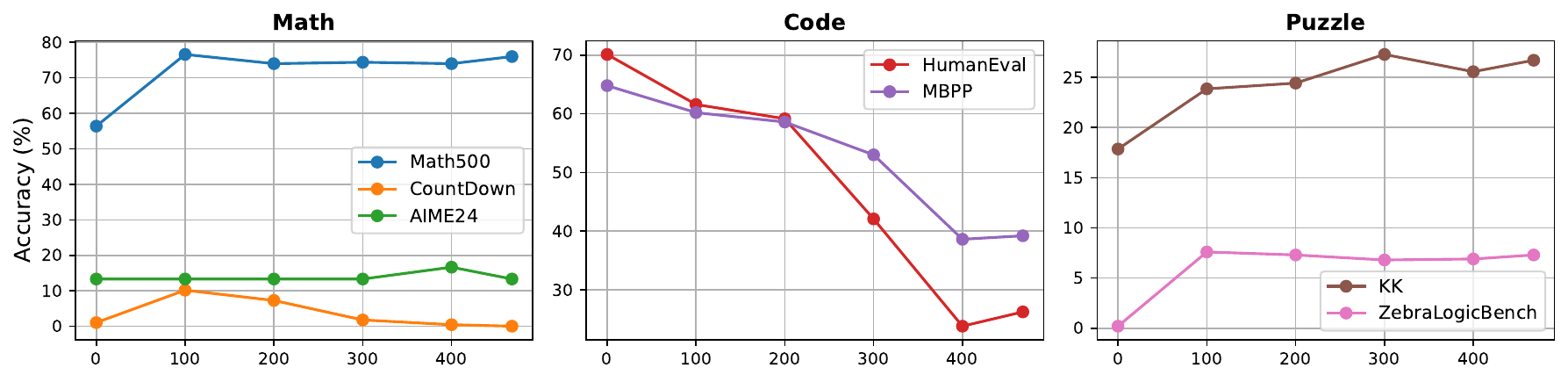}
    \caption{Base model’s detailed performance on DeepscaleR.}
    \label{single_deepscaler1}
\end{figure}

\begin{figure}[h]
    \centering
    \includegraphics[width=0.95\textwidth]{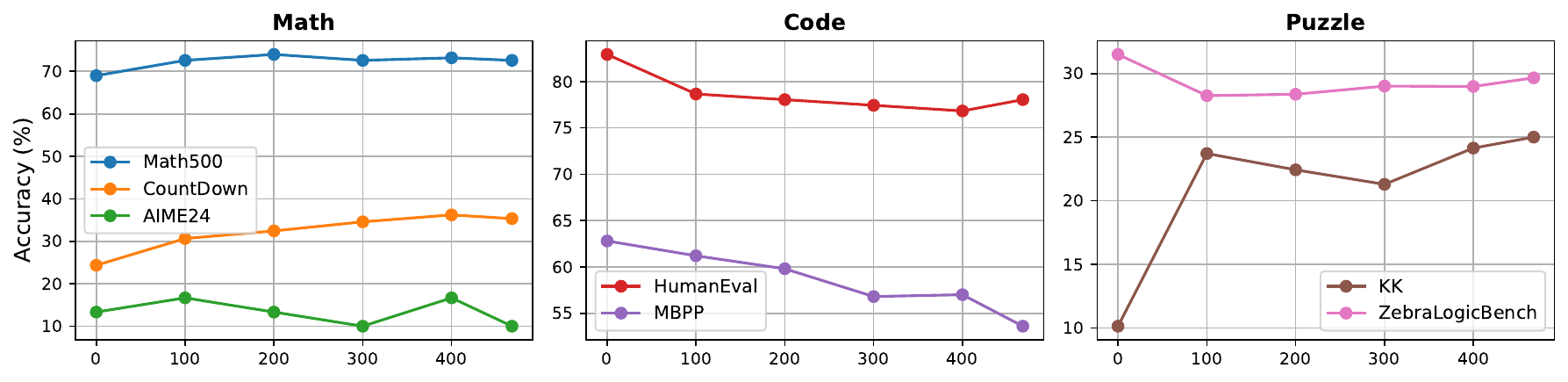}
    \caption{Instruct model’s detailed performance on DeepscaleR.}
    \label{single_deepscaler2}
\end{figure}

    \item \textbf{CountDown:} Figures~\ref{single_CountDown_base} and ~\ref{single_CountDown_inst} present the training dynamics on the CountDown dataset. Although the model initially underperforms on CountDown, domain-specific training significantly improves its results. Meanwhile, CountDown training degrades the base model’s coding ability, whereas the instruct model’s coding performance remains relatively stable, indirectly demonstrating the positive impact of SFT on the robustness of RL training.
\vspace{-1mm}
\begin{figure}[h]
    \centering
    \includegraphics[width=0.95\textwidth]{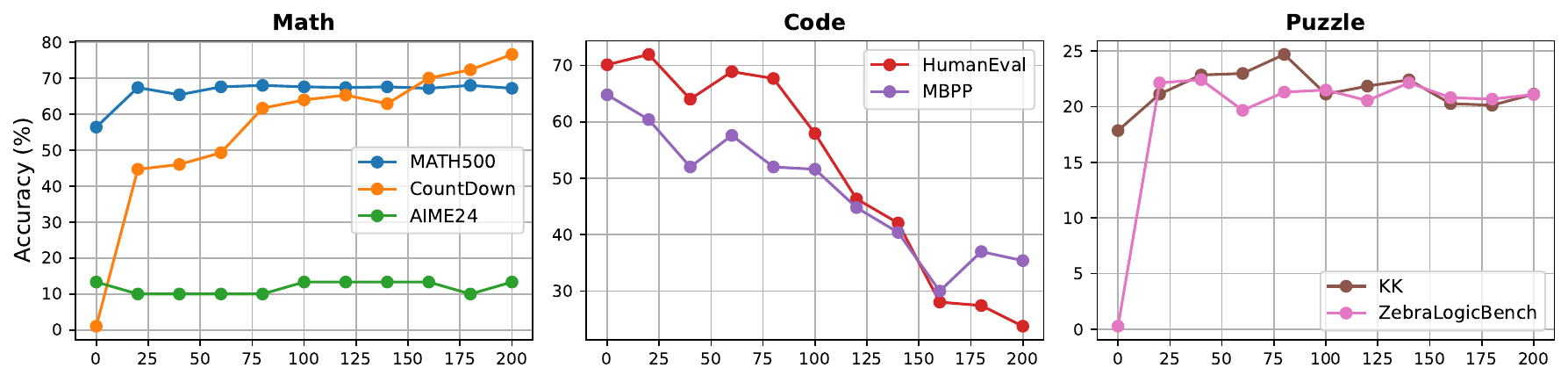}
    \caption{Base model’s detailed performance on CountDown.}
    \label{single_CountDown_base}
\end{figure}

\begin{figure}[h]
    \centering
    \includegraphics[width=0.95\textwidth]{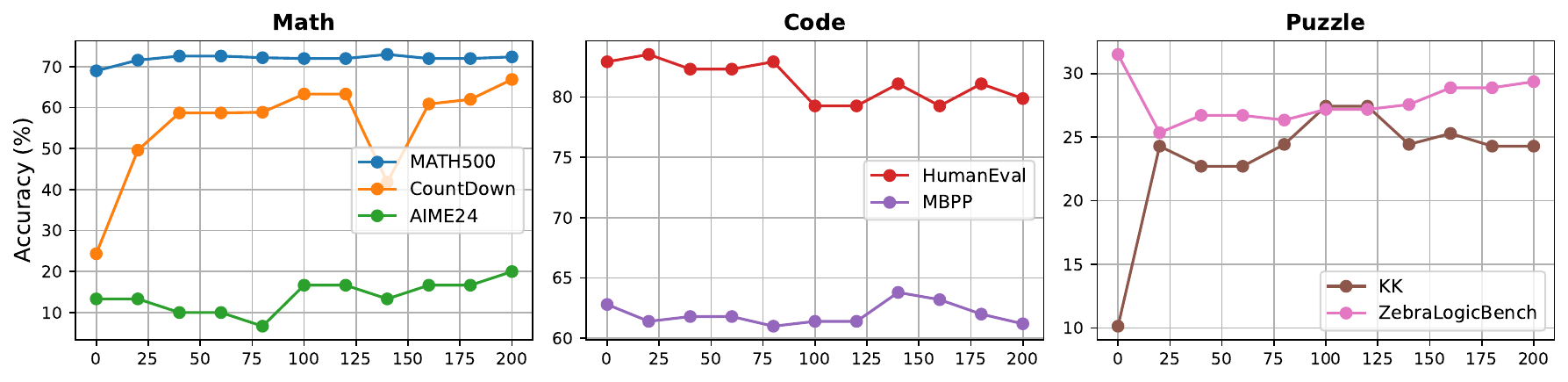}
    \caption{Instruct model’s detailed performance on CountDown.}
    \label{single_CountDown_inst}
    \vspace{-0.6cm}
\end{figure}

    \item \textbf{CodeR1:} Figures~\ref{single_code} and ~\ref{single_code_instruct} show the training dynamics on the CodeR1 dataset. While code data substantially improves the models’ coding reasoning ability, it also introduces negative effects on mathematical reasoning, with the impact being more pronounced for the base model.

\begin{figure}[h]
    \centering
    \includegraphics[width=0.95\textwidth]{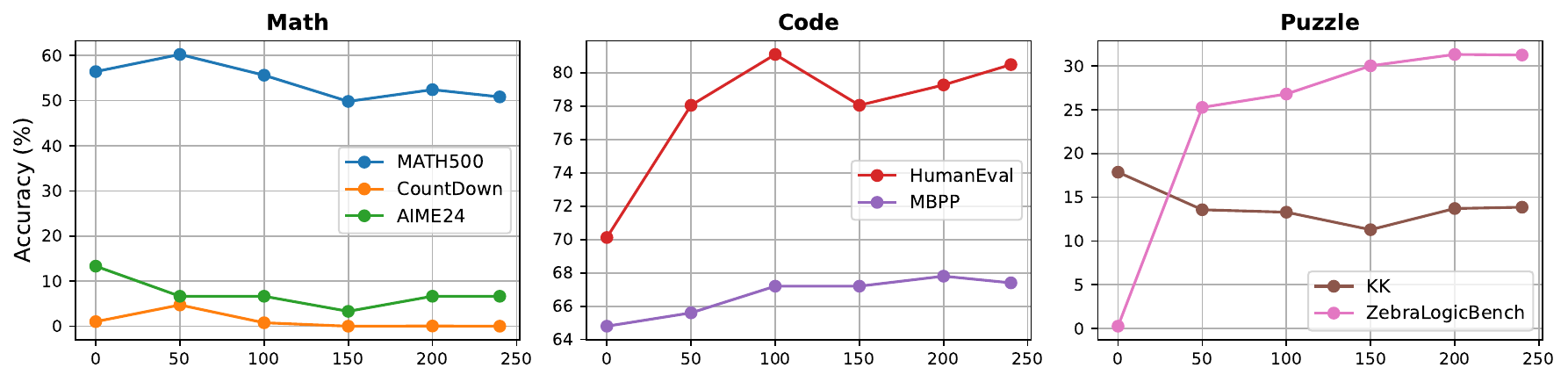}
    \caption{Base model’s detailed performance on CodeR1.}
    \label{single_code}
    \vspace{-0.6cm}
\end{figure}

\begin{figure}[h]
    \centering
    \includegraphics[width=0.95\textwidth]{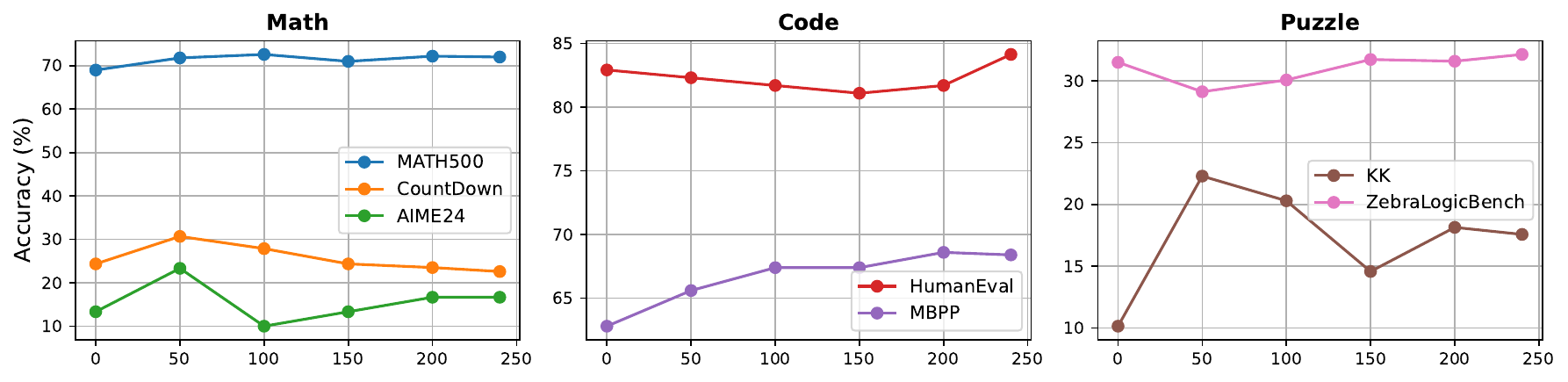}
    \caption{Instruct model’s detailed performance on CodeR1.}
    \label{single_code_instruct}
\end{figure}

    \item \textbf{Knights-and-Knaves:} Figures~\ref{single_kk} and ~\ref{single_kk_instruct} show the effects of training on the KK dataset. Targeted training significantly boosts KK performance, with this improvement in logical reasoning generalizing well to mathematical tasks but negatively impacting coding ability.

    \begin{figure}[h!]
        \centering
        \includegraphics[width=0.95\textwidth]{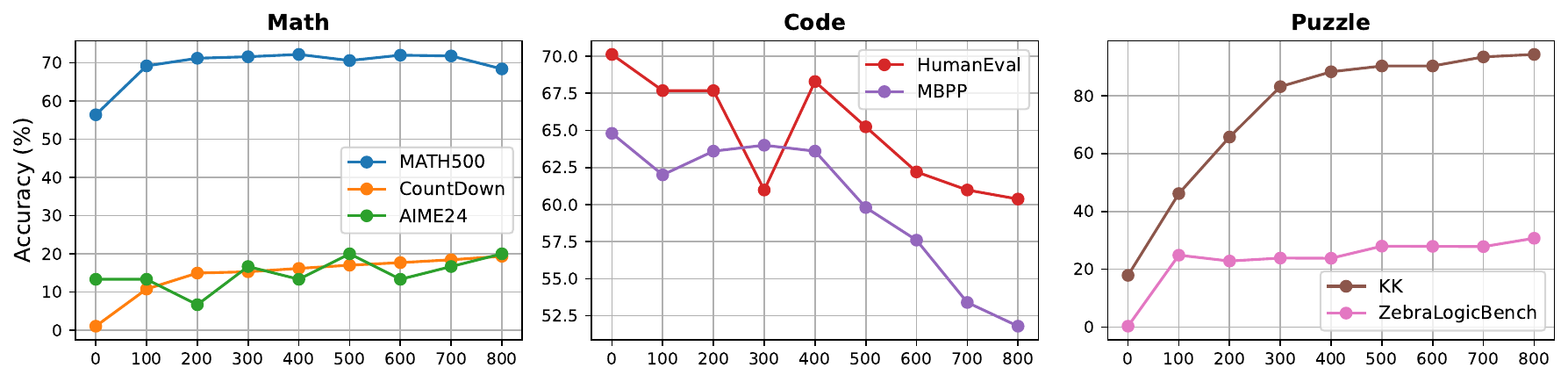}
        \caption{Base model’s detailed performance on KK.}
        \label{single_kk}
    \end{figure}

    \begin{figure}[h!]
        \centering
        \includegraphics[width=0.95\textwidth]{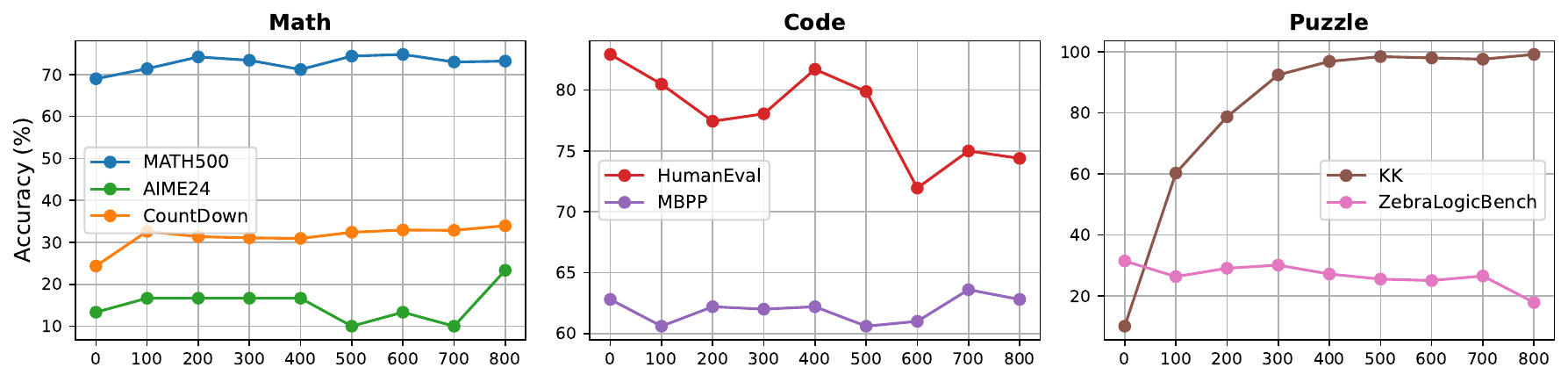}
        \caption{Instruct model’s detailed performance on KK.}
        \label{single_kk_instruct}
    \end{figure}
    
    \item \textbf{Logic Puzzle Baron:} Figures~\ref{single_zebra_base} and ~\ref{single_zebra_inst} show the effects of training on the LPB dataset. Similar to the KK setting, the model’s performance on the ZebraLogicBench improves rapidly with targeted training.
\begin{figure}[h!]
    \centering
    \includegraphics[width=0.95\textwidth]{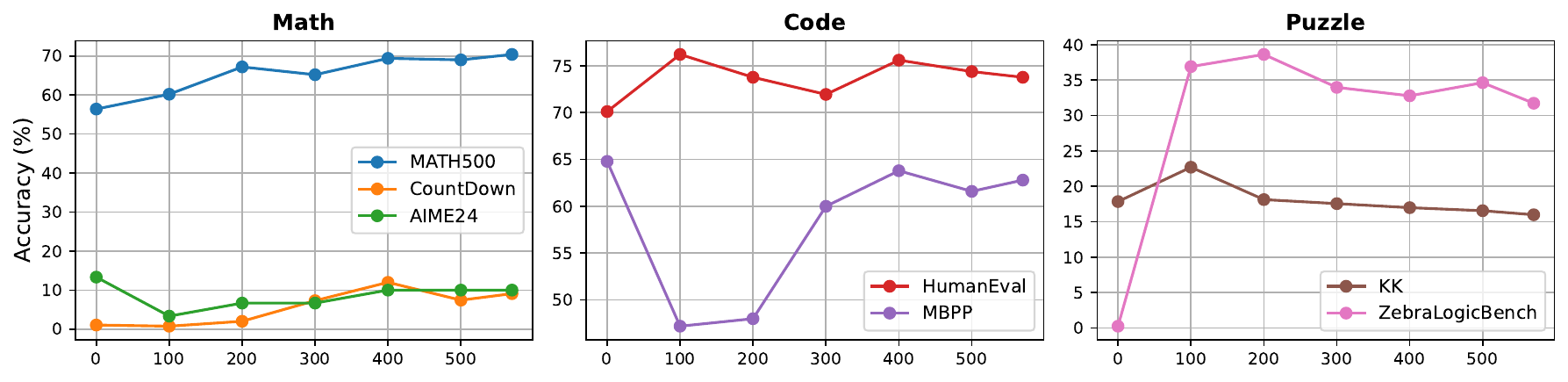}
    \caption{Base model’s detailed performance on LPB.}
    \vspace{-0.6cm}
    \label{single_zebra_base}
\end{figure}

\begin{figure}[h!]
    \centering
    \includegraphics[width=0.95\textwidth]{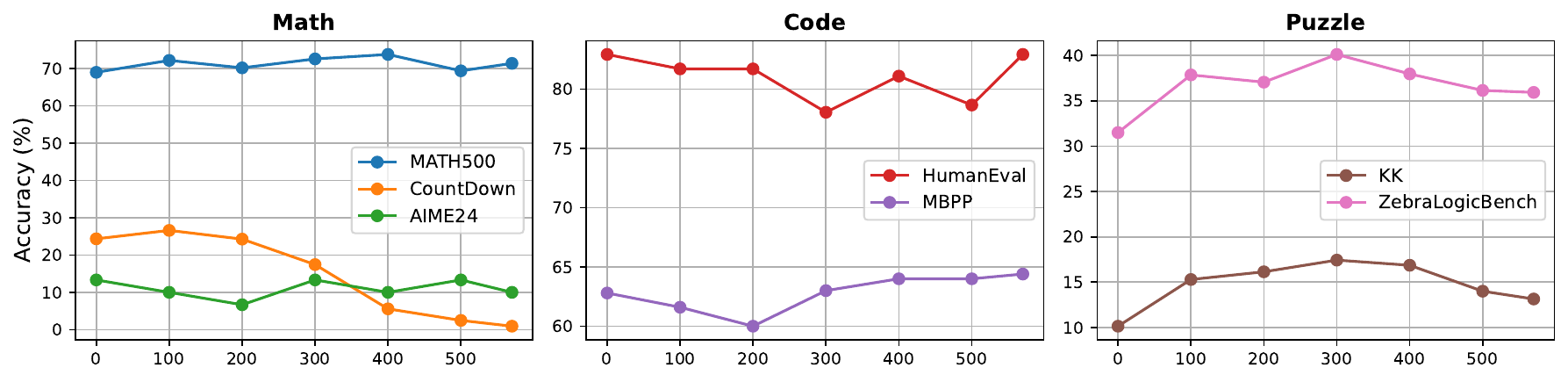}
    \caption{Instruct model’s detailed performance on LPB.}
    \vspace{-0.6cm}
    \label{single_zebra_inst}
\end{figure}

\end{itemize}

\section{Detailed Performance Results for Cross-Domain Composition}
\begin{itemize}
    \item \textbf{Puzzle + Math}: As shown in Figure~\ref{puzzle_math}, under this setting, the model's performance on math and puzzle tasks shows a steady improvement. However, for HumanEval, there is a slight improvement followed by a significant decline, which also demonstrates the catastrophic forgetting phenomenon in RLVR.
    \begin{figure}[h!]
    \centering
    \includegraphics[width=0.95\textwidth]{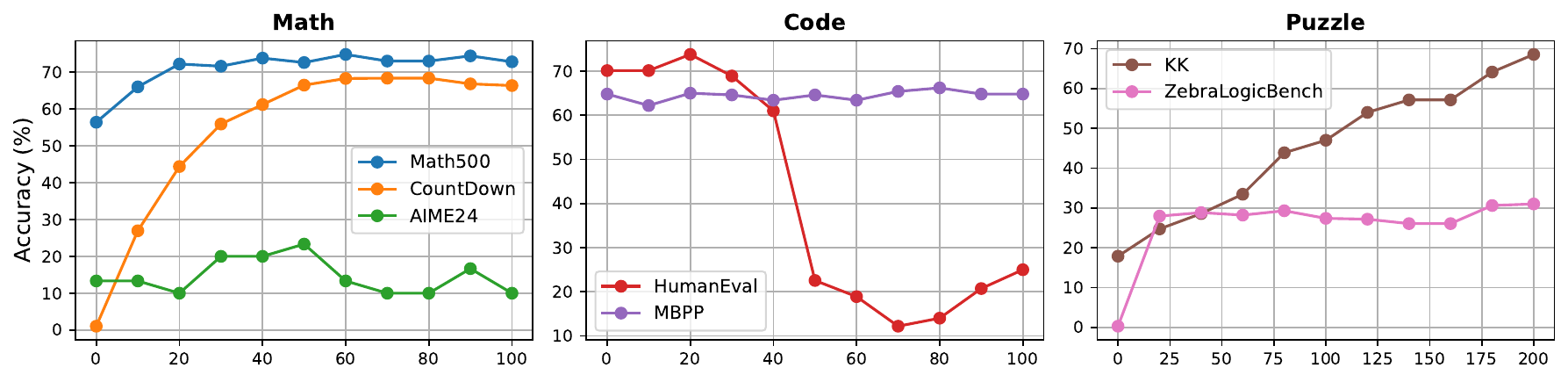}
    \caption{Base model’s detailed performance on Puzzle + Math domain data.}

    \label{puzzle_math}
\end{figure}

    \item \textbf{Math + Code}: As shown in Figure~\ref{fig_math_code}, the model exhibits stable improvements across all aspects in this setting, both in-domain for math and code tasks. Additionally, the extra math data further enhances performance on code tasks. The model also generalizes to unseen puzzle-solving capabilities.
\begin{figure}[h!]
    \centering
    \includegraphics[width=0.95\textwidth]{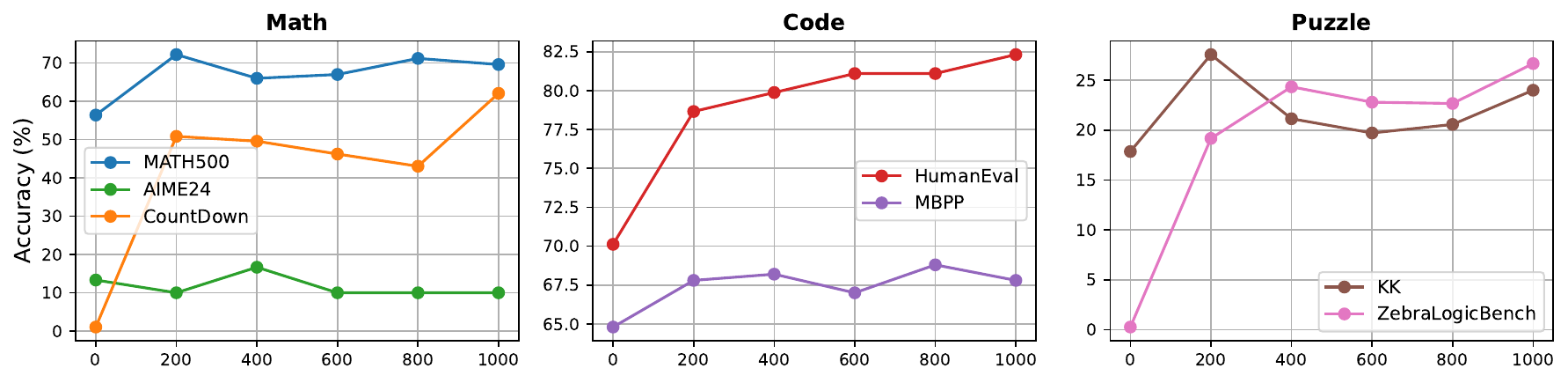}
    \caption{Base model’s detailed performance on Math + Code domain data.}
    \label{fig_math_code}
\end{figure}

\item \textbf{Puzzle + Code}: As shown in Figure~\ref{fig_puzzle_code}, although code performance shows gradual improvement, the in-domain puzzle performance does not exhibit significant improvement compared to other settings, indicating that code data has an additional negative impact on puzzle performance.
\begin{figure}[h!]
    \centering
    \includegraphics[width=0.95\textwidth]{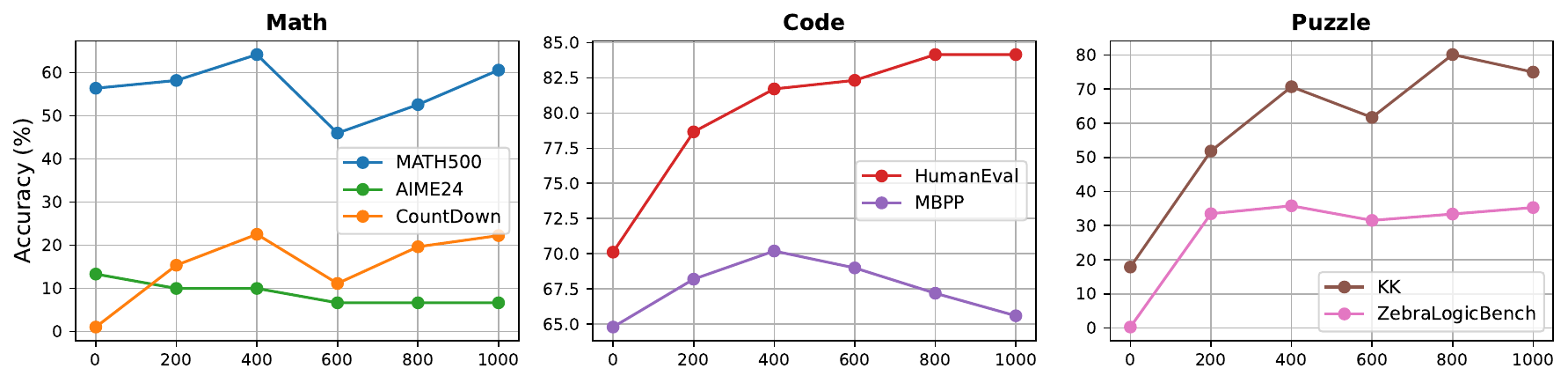}
    \caption{Base model’s detailed performance on Puzzle + Code domain data.}
    \label{fig_puzzle_code}
\end{figure}

\end{itemize}

In addition, we have summarized the specific performance of all cross-domain combinations in Table~\ref{tab:two_domain_comb}:

\begin{table}[h!]
\centering
\caption{Model performance during cross-domain data composition.}
\label{tab:two_domain_comb}
\resizebox{\textwidth}{!}
{
\begin{tabular}{lccccccccccc}
\toprule
\multirow{2}{*}{\textbf{Training Data}}  & \multicolumn{4}{c}{\textbf{Math}} & \multicolumn{3}{c}{\textbf{Code}} & \multicolumn{3}{c}{\textbf{Puzzle}} & \textbf{ALL}\\
\cmidrule(r){2-5} \cmidrule(r){6-8} \cmidrule(r){9-11} \cmidrule(r){12-12}
& MATH500 & CountDown & AIME24 & Avg. & HumanEval & MBPP & Avg. & KK & Zebra & Avg.& Avg.\\
\midrule
Base & 56.40 & 1.05 & 10.00 & 22.48 & 70.12 & 64.80 & 67.46 & 17.86 & 0.27 & 9.07 &31.50\\
\addlinespace[2pt]
\hdashline
\addlinespace[2pt]
Math & 72.00 & 53.77 & 16.67 & 47.48 & 66.46 & 62.00 & 64.23 & 26.42 & 18.42 & 22.42 &45.11\\
Puzzle & 67.60 & 10.81 & 10.00 & 29.47 & 78.70 & 64.00 & 71.35 & 89.29 & 34.66 & 61.98 &50.72\\
Code & 50.80 & 0.04 & 6.67 & 19.17 & 80.49 & 67.40 & 73.95 & 13.85 & 31.24 & 22.55 &35.78\\
\addlinespace[2pt]
\hdashline
\addlinespace[2pt]
Math + Puzzle & 72.80 & 66.35 & 10.00 & 49.72 & 25.00 & 64.80 & 44.90 & 68.57 & 30.99 & 49.78 &48.36\\
Puzzle + Code & 60.60 & 22.25 & 13.33 & 32.06 & 84.15 & 65.60 & 74.88 & 75.00 & 35.30 & 55.15 & 50.89 \\
Math + Code &  69.60 & 62.07 & 10.00 & 47.22 & 82.32 & 67.80 & 75.06 & 24.00 & 26.67 & 25.34 & 48.92\\
\addlinespace[2pt]
\hdashline
\addlinespace[2pt]
Math + Puzzle + Code & 73.60 & 52.33 & 23.33 & 49.75&  78.66 & 68.60 & 73.63 & 68.14 & 31.31 & 49.73 & 56.57\\
\bottomrule
\end{tabular}
}
\end{table}

\section{Prompt Formats for Each Benchmark Evaluation}
\label{prompt}

\begin{figure*}[h!]
    \centering
    \begin{exmp}{MATH500}{parallel}
        \small
        \setstretch{1.4}  
Now the user asks you to solve a math problem. After thinking, when you finally reach a conclusion, clearly state the answer within \textless answer\textgreater\ \textless/answer\textgreater\ tags. i.e., \textless answer\textgreater\ (\textbackslash boxed\{\}) \textless/answer\textgreater. \{\bbb{problem}\}\textbackslash n
      
    \end{exmp}
\end{figure*}

\begin{figure*}[htbp]
    \centering
    \begin{exmp}{AIME24}{parallel}
        \small
        \setstretch{1.4}  
Now the user asks you to solve a math problem. After thinking, when you finally reach a conclusion, clearly state the answer within \textless answer\textgreater\ \textless/answer\textgreater\ tags. i.e., \textless answer\textgreater\ (\textbackslash boxed\{\}) \textless/answer\textgreater. \{\bbb{question}\}\textbackslash n
      
    \end{exmp}
\end{figure*}

\begin{figure*}[htbp]
    \centering
    \begin{exmp}{CountDown}{parallel}
        \small
        \setstretch{1.4}  
Now the user asks you to solve a math problem. After thinking, when you finally reach a conclusion, clearly state the answer within \textless answer\textgreater\ \textless/answer\textgreater\ tags. i.e., \textless answer\textgreater\ \textless/answer\textgreater. Using the numbers \{\bbb{numbers}\}, create an equation that equals \{\bbb{target}\}. You can use basic arithmetic operations (+, -, *, /) and each number can only be used once. Show your work in \textless think\textgreater\ \textless/think\textgreater\ tags. And return the final answer in \textless answer\textgreater\ \textless/answer\textgreater\ tags, for example \textless answer\textgreater\ (1 + 2) / 3 \textless/answer\textgreater.
      
    \end{exmp}
\end{figure*}

\begin{figure*}[h!]
    \centering
    \begin{exmp}{HumanEval}{parallel}
        \small
        \setstretch{1.4}  
Now the user asks you to solve a code problem. After thinking, when you finally reach a conclusion, clearly state the answer within \textless answer\textgreater\ \textless/answer\textgreater\ tags. i.e., \textless answer\textgreater\ \textless/answer\textgreater. Complete the following python code:\textbackslash n\{\bbb{prompt}\}\textbackslash n
      
    \end{exmp}
\end{figure*}

\begin{figure*}[h!]
    \centering
    \begin{exmp}{Knights-and-Knaves (KK)}{parallel}
        \small
        \setstretch{1.4}  
Now the user asks you to solve a logical reasoning problem. After thinking, when you finally reach a conclusion, clearly state the identity of each character within \textless answer\textgreater\ \textless/answer\textgreater\ tags. List the identity of each person one by one, for example, \textless answer\textgreater\ (1) Zoey is a knight (2) Oliver is a knight (3)... \textless/answer\textgreater.
New question:
\{\bbb{prompt}\}
      
    \end{exmp}
\end{figure*}

\begin{figure*}[htbp]
    \centering
    \begin{exmp}{MBPP}{parallel}
        \small
        \setstretch{1.4}  
You are an expert Python programmer, and here is your task: Write a function to find the similar elements from the given two tuple lists. Your code should pass these tests:

assert similar\_elements((3, 4, 5, 6),(5, 7, 4, 10))==(4, 5)\\
assert similar\_elements((1, 2, 3, 4),(5, 4, 3, 7)) == (3, 4)\\
assert similar\_elements((11, 12, 14, 13),(17, 15, 14, 13)) == (13, 14)\\

[BEGIN]\\
\quad 'def similar\_elements(test\_tup1, test\_tup2):

    \qquad res = tuple(set(test\_tup1) \& set(test\_tup2))
    
   \qquad return (res)'\\
\quad [DONE] \\

You are an expert Python programmer, and here is your task: Write a python function to identify non-prime numbers. Your code should pass these tests:

assert is\_not\_prime(2) == False\\
assert is\_not\_prime(10) == True\\
assert is\_not\_prime(35) == True\\

[BEGIN]\\
\quad 'import math\\
def is\_not\_prime(n):

  \qquad   result = False
  
  \qquad   for i in range\{ 2,int(math.sqrt(n)) + 1 \}:
  
     \qquad \qquad    if n \% i == 0:
     
         \qquad \qquad \qquad    result = True
         
   \qquad  return result
'\\
\quad [DONE]\\

You are an expert Python programmer, and here is your task: Write a function to find the largest integers from a given list of numbers using heap queue algorithm. Your code should pass these tests:

assert heap\_queue\_largest( [25, 35, 22, 85, 14, 65, 75, 22, 58],3\} == (85, 75, 65)\\
assert heap\_queue\_largest( [25, 35, 22, 85, 14, 65, 75, 22, 58],2) == (85, 75)\\
assert heap\_queue\_largest( [25, 35, 22, 85, 14, 65, 75, 22, 58],5) == (85, 75, 65, 58, 35)\\

[BEGIN]\\
\quad ' import heapq as hq\\
def heap\_queue\_largest(nums,n):

  \qquad   largest\_nums = hq.nlargest(n, nums)
  
  \qquad   return largest\_nums
'\\
\quad [DONE]\\

You are an expert Python programmer, and here is your task: \{\bbb{text}\} Your code should pass these tests:

\{\bbb{test\_list}\}

    \end{exmp}
\end{figure*}

\begin{figure*}[htbp]
    \centering
    \begin{exmp}{ZebraLogicBench (Zebra)}{parallel}
        \small
        \setstretch{1.4}  
\# Example Puzzle 

There are 3 houses, numbered 1 to 3 from left to right, as seen from across the street. Each house is occupied by a different person. Each house has a unique attribute for each of the following characteristics:
 
 - Each person has a unique name: `Peter', `Eric', `Arnold'.
 
 - Each person has a unique favorite drink: `tea', `water', `milk'

\#\# Clues for the Example Puzzle

1. Peter is in the second house.

2. Arnold is directly left of the one who only drinks water.

3. The one who only drinks water is directly left of the person who likes milk.

\#\# Answer to the Example Puzzle

\{  

\qquad"reasoning": "Given Clue 1, we know Peter is in House 2. According to Clue 2, Arnold is directly left of the one who only drinks water. The person in House 3 cannot be on the left of anyone, so Arnold must be in House 1. Thus, Peter drinks water, and Eric lives in House 3. Then, according to Clue 3, Eric drinks milk. Therefore, Arnold drinks tea.", 

\qquad\qquad"solution": \{

        \qquad\qquad\qquad"House 1": \{ 
        
          \qquad\qquad\qquad\qquad  "Name": "Arnold",
          
           \qquad\qquad\qquad\qquad "Drink": "tea"  
           
      \qquad\qquad\qquad  \}, 
      
      \qquad\qquad\qquad  "House 2": \{ 
      
        \qquad\qquad\qquad\qquad    "Name": "Peter",
        
        \qquad\qquad\qquad\qquad    "Drink": "water" 
            
       \qquad\qquad\qquad \},
       
     \qquad\qquad\qquad   "House 3": \{ 
        
   \qquad\qquad\qquad\qquad  "Name": "Eric",
            
      \qquad\qquad\qquad\qquad  "Drink": "milk"
        
       \qquad\qquad\qquad \}  
       
  \qquad\qquad  \}  
    
\}

\# Puzzle to Solve 

\{\bbb{puzzle}\}

\# Instruction

Now please solve the above puzzle. Present your reasoning and solution in the following json format:

\{\bbb{json\_template}\}
      
    \end{exmp}
\end{figure*}

\end{document}